\PassOptionsToPackage{unicode}{hyperref}
\PassOptionsToPackage{hyphens}{url}
\PassOptionsToPackage{dvipsnames,svgnames,x11names}{xcolor}
\documentclass[
  12pt,
]{article}
\usepackage{xcolor}
\usepackage[lmargin=1in,rmargin=1in,tmargin=1in,bmargin=1in]{geometry}
\usepackage{amsmath,amssymb}
\usepackage{iftex}
\ifPDFTeX
  \usepackage[T1]{fontenc}
  \usepackage[utf8]{inputenc}
  \usepackage{textcomp} % provide euro and other symbols
\else % if luatex or xetex
  \usepackage{unicode-math} % this also loads fontspec
  \defaultfontfeatures{Scale=MatchLowercase}
  \defaultfontfeatures[\rmfamily]{Ligatures=TeX,Scale=1}
\fi
\usepackage{lmodern}
\ifPDFTeX\else
\fi
\IfFileExists{upquote.sty}{\usepackage{upquote}}{}
\IfFileExists{microtype.sty}{% use microtype if available
  \usepackage[]{microtype}
  \UseMicrotypeSet[protrusion]{basicmath} % disable protrusion for tt fonts
}{}
\usepackage{setspace}
\makeatletter
\@ifundefined{KOMAClassName}{% if non-KOMA class
  \IfFileExists{parskip.sty}{%
    \usepackage{parskip}
  }{% else
    \setlength{\parindent}{0pt}
    \setlength{\parskip}{6pt plus 2pt minus 1pt}}
}{% if KOMA class
  \KOMAoptions{parskip=half}}
\makeatother
\makeatletter
\ifx\paragraph\undefined\else
  \let\oldparagraph\paragraph
  \renewcommand{\paragraph}{
    \@ifstar
      \xxxParagraphStar
      \xxxParagraphNoStar
  }
  \newcommand{\xxxParagraphStar}[1]{\oldparagraph*{#1}\mbox{}}
  \newcommand{\xxxParagraphNoStar}[1]{\oldparagraph{#1}\mbox{}}
\fi
\ifx\subparagraph\undefined\else
  \let\oldsubparagraph\subparagraph
  \renewcommand{\subparagraph}{
    \@ifstar
      \xxxSubParagraphStar
      \xxxSubParagraphNoStar
  }
  \newcommand{\xxxSubParagraphStar}[1]{\oldsubparagraph*{#1}\mbox{}}
  \newcommand{\xxxSubParagraphNoStar}[1]{\oldsubparagraph{#1}\mbox{}}
\fi
\makeatother

\usepackage{longtable,booktabs,array}
\usepackage{calc} % for calculating minipage widths
\usepackage{etoolbox}
\makeatletter
\patchcmd\longtable{\par}{\if@noskipsec\mbox{}\fi\par}{}{}
\makeatother
\IfFileExists{footnotehyper.sty}{\usepackage{footnotehyper}}{\usepackage{footnote}}
\makesavenoteenv{longtable}
\usepackage{graphicx}
\makeatletter
\newsavebox\pandoc@box
\newcommand*\pandocbounded[1]{% scales image to fit in text height/width
  \sbox\pandoc@box{#1}%
  \Gscale@div\@tempa{\textheight}{\dimexpr\ht\pandoc@box+\dp\pandoc@box\relax}%
  \Gscale@div\@tempb{\linewidth}{\wd\pandoc@box}%
  \ifdim\@tempb\p@<\@tempa\p@\let\@tempa\@tempb\fi% select the smaller of both
  \ifdim\@tempa\p@<\p@\scalebox{\@tempa}{\usebox\pandoc@box}%
  \else\usebox{\pandoc@box}%
  \fi%
}
\def\fps@figure{htbp}
\makeatother

\NewDocumentCommand\citeproctext{}{}

\makeatletter
 \let\@cite@ofmt\@firstofone
 \def\@biblabel#1{}
 \def\@cite#1#2{{#1\if@tempswa , #2\fi}}
\makeatother
\newlength{\cslhangindent}
\newlength{\csllabelwidth}
\newenvironment{CSLReferences}[2] % #1 hanging-indent, #2 entry-spacing
 {\begin{list}{}{%
  \setlength{\itemindent}{0pt}
  \setlength{\leftmargin}{0pt}
  \setlength{\parsep}{0pt}
  \ifodd #1
   \setlength{\leftmargin}{\cslhangindent}
   \setlength{\itemindent}{-1\cslhangindent}
  \fi
  \setlength{\itemsep}{#2\baselineskip}}}
 {\end{list}}
\usepackage{calc}

\providecommand{\tightlist}{%
  \setlength{\itemsep}{0pt}\setlength{\parskip}{0pt}}

\usepackage{rotating}
\usepackage{titling}  % Provides \subtitle command
\usepackage{caption}
\usepackage{newfloat}
\DeclareFloatingEnvironment{prompt}

\newcommand{\promptref}[1]{\hyperref[{#1}]{Prompt~\ref*{#1}}}

\usepackage{listings}
\usepackage[most]{tcolorbox}

\lstdefinelanguage{Jinja}{
  basicstyle=\ttfamily,
  sensitive=true,
  morestring=[b]",
  moredelim=[s][\color{blue}]{\{\{}{\}\}},
  moredelim=[s][\color{purple}]{\{\%}{\%\}},
}

\newtcolorbox[list inside=prompt,auto counter,number within=section]{promptbox}[1][]{
    colbacktitle=black!60,
    fonttitle=\small,
    coltitle=white,
    fontupper=\footnotesize,
    boxsep=3pt,
    left=0pt,
    right=0pt,
    top=0pt,
    bottom=0pt,
    boxrule=1pt,
    #1
}

\usepackage{authblk}
\usepackage{titling}
\usepackage{booktabs}
\usepackage{adjustbox}
\usepackage[labelfont=bf]{caption}
\usepackage{multirow}
\usepackage{pdflscape}
\usepackage{afterpage}
\usepackage{newunicodechar}
\usepackage{titletoc}
\newunicodechar{ρ}{$\rho$}
\newunicodechar{Δ}{$\Delta$}
\makeatletter
\@ifpackageloaded{caption}{}{\usepackage{caption}}
\AtBeginDocument{%
\ifdefined\contentsname
  \renewcommand*\contentsname{Table of contents}
\else
  \newcommand\contentsname{Table of contents}
\fi
\ifdefined\listfigurename
  \renewcommand*\listfigurename{List of Figures}
\else
  \newcommand\listfigurename{List of Figures}
\fi
\ifdefined\listtablename
  \renewcommand*\listtablename{List of Tables}
\else
  \newcommand\listtablename{List of Tables}
\fi
\ifdefined\figurename
  \renewcommand*\figurename{Figure}
\else
  \newcommand\figurename{Figure}
\fi
\ifdefined\tablename
  \renewcommand*\tablename{Table}
\else
  \newcommand\tablename{Table}
\fi
}
\@ifpackageloaded{float}{}{\usepackage{float}}
\floatstyle{ruled}
\@ifundefined{c@chapter}{\newfloat{codelisting}{h}{lop}}{\newfloat{codelisting}{h}{lop}[chapter]}
\floatname{codelisting}{Listing}
\makeatother
\makeatletter
\@ifpackageloaded{caption}{}{\usepackage{caption}}
\@ifpackageloaded{subcaption}{}{\usepackage{subcaption}}
\makeatother
\usepackage{bookmark}
\IfFileExists{xurl.sty}{\usepackage{xurl}}{} % add URL line breaks if available
\hypersetup{
  pdftitle={Computational emotion analysis with multimodal LLMs},
  pdfauthor={Hauke Licht},
  colorlinks=true,
  linkcolor={black},
  filecolor={Maroon},
  citecolor={black},
  urlcolor={black},
  pdfcreator={LaTeX via pandoc}}

\title{%
    Computational emotion analysis with multimodal LLMs
    \\\Large Current evidence on an emerging methodological opportunity%
}

  \author{Hauke Licht\thanks{\href{mailto:hauke.licht@uibk.ac.at}{hauke.licht@uibk.ac.at}}}
            \affil{%
                  University of Innsbruck
                                      }
      
\date{April 3, 2026}
\begin{document}
\maketitle
\begin{abstract}
\noindent Research increasingly leverages audio-visual materials to analyze emotions in political communication.
Multimodal large language models (mLLMs) promise to enable such analyses through in-context learning.
However, we lack systematic evidence on whether current mLLMs can reliably measure emotions in real-world political settings.
This paper closes this gap by evaluating open- and closed-weights mLLMs available as of early 2026 in video-based emotional arousal measurement using two complementary human-labeled datasets: speech actor recordings created under laboratory conditions and real-world parliamentary debates.
I find a critical lab-vs-field performance gap.
In videos created under laboratory conditions, the examined mLLMs arousal scores approach human-level reliability.
However, in parliamentary debate recordings, all examined models' arousal scores correlate at best moderately with average human ratings.
Moreover, in each dataset, all but one of the examined mLLMs exhibit systematic gender-differential bias, consistently underestimating arousal more for male than for female speakers, resulting in a net-positive intensity bias.
These findings reveal important limitations of current mLLMs for real-world political video analysis and establish a rigorous evaluation framework for tracking future developments.

\end{abstract}

\setstretch{1.5}
\begin{center}
Word count: 7156
\end{center}
% % TODO: add for journal submission

\thispagestyle{empty}
\clearpage
\setcounter{page}{1}
\setlength{\parskip}{0pt}
\setlength{\parindent}{20pt}

\section{Introduction}\label{introduction}

Large pre-trained generative language models (LLMs) are revolutionizing computational applications in political science.
\emph{In-context learning} (Brown et al. 2020), in particular, where a model receives task instructions and generates a response that is treated as an annotation, has been adopted widely for text classification (e.g., Gilardi, Alizadeh, and Kubli 2023), scoring (O'Hagan and Schein 2023; Mens and Gallego 2024; Licht et al. 2025; Benoit et al. 2025), and extraction (Kasner et al. 2025; Stuhler, Ton, and Ollion 2025).

However, political communication involves not only \emph{what} is being said but also \emph{how} it is said (e.g., Dietrich, Hayes, and O'brien 2019; Boussalis et al. 2021; Cochrane et al. 2022; Rittmann, Ringwald, and Nyhuis 2025).
Further, video platforms like YouTube and TikTok are increasingly important for political messaging.
Focusing only on text therefore limits our understanding of political communication and calls for audio-visual analyses.

While recent work has validated LLMs for text-based tasks, and others have proposed specialized models for emotion analysis in videos, no research has systematically evaluated whether \emph{multimodal} LLMs (mLLMs) can reliably measure emotions in political videos.
This gap is critical because mLLMs can process text, audio, images, and videos while following textual instructions, promising conceptually guided multimodal measurements with minimal labeled data requirements through in-context learning.

This paper investigates the in-context learning capabilities of mLLMs for emotion analysis, focusing on measuring emotional arousal in speech.
Emotional arousal manifests in audio-visual cues such as vocal tone, facial expressions, and body language (cf. Dietrich, Hayes, and O'brien 2019; Boussalis et al. 2021; Cochrane et al. 2022; Rittmann, Ringwald, and Nyhuis 2025), making it well-suited for multimodal analysis.

I evaluate the first available mLLMs capable of processing temporally aligned audio-visual input as of early 2026, focusing on open-weights models in the Qwen Omni (released mid 2025; Xu, Guo, He, et al. (2025); Xu, Guo, Hu, et al. (2025)) and TowerVideo (released in October 2025; Viveiros et al. (2025)) series and Google's closed-weights Gemini Flash models (released between late 2024 and 2025; Google (2024); Google (2025a)).
I rely on two complementary datasets with high-quality human-annotated emotional arousal ratings for evaluation.
The first dataset, RAVDESS (Livingstone and Russo 2018), contains short recordings of speech actors created under laboratory conditions and thus presents a ``most likely'' scenario for reliable video-based emotion intensity scoring.
The second dataset contains segments of speeches recorded in the Canadian \emph{House of Commons} (Cochrane et al. 2022), providing a more challenging yet realistic test case for evaluating mLLMs in real-world political contexts.

My findings document a critical lab-vs-field performance gap.
Under ideal laboratory conditions, Gemini Flash and Qwen Omni models approach human-level measurement reliability.
However, when deployed in real-world political debates, all examined mLLMs fail to deliver on this promise, as their arousal measurements correlate at best moderately with average human ratings.
Moreover, in each dataset, all but one examined mLLMs exhibit systematic gender-differential bias, underestimating arousal more for male than female speakers -- a pattern that may compound gender bias already present in human annotations.
And even in sentiment scoring -- where mLLMs perform overall better than in arousal rating -- their video-based performance lags markedly behind text-based scoring.
Overall, these findings highlight both the potential and the limitations of the examined mLLMs at this stage of their development for video-based emotion analysis in political communication research.

This paper makes three contributions.
First, it provides the first systematic evaluation of mLLMs for video-based emotion measurement in political contexts.
Second, it identifies a critical lab-to-field performance gap and demonstrates that neither model scale nor denoising strategies close this gap.
Third, its replicable evaluation framework and evidence set a temporal anchor, enabling systematic comparison of future model releases.

\section{Background}\label{sec-background}

\subsection{Emotions in political communication}\label{emotions-in-political-communication}

Emotions are central in politics because how elites use and display them can shape audiences' perceptions and mobilization.
Classical models of political decision-making highlight that affective responses are integral to political cognition and behavior (Bakker, Schumacher, and Rooduijn 2021; Lodge and Taber 2005, 2013).
Because politicians express emotions in what they say, how they say it, and how they look when saying it, this influence runs through verbal, vocal, and visual channels of political communication.
For example, vocal pitch, facial expressions or gestures can affect candidate evaluations by influencing perceptions of candidates' authority and leadership ability (Druckman 2003; Dumitrescu, Gidengil, and Stolle 2015; Klofstad, Anderson, and Peters 2012; Anderson and Klofstad 2012; Klofstad 2016; Cinar and Kıbrıs 2024), emotions can drive political engagement (Brader 2005; Valentino et al. 2011), and physiological measures of arousal have been shown to predict subjects' attitude change (Bakker, Schumacher, and Rooduijn 2021).

For politicians, the display of emotions serves both expressive and strategic purposes.
On the one hand, elites may display emotions unintentionally, arising from physiological activation that speakers cannot easily suppress (Dietrich, Enos, and Sen 2019; Dietrich, Hayes, and O'brien 2019).
On the other hand, elites' emotional expression may also follow a strategic calculus.
Speakers may strategically modulate tone, posture, and facial expression to emphasize points, signal opposition and conflict, or connect with audiences (Osnabrügge, Hobolt, and Rodon 2021; Rask and Hjorth 2025; Arnold and Küpfer 2025; Rittmann, Nyhuis, and Ringwald 2025).

\subsection{Computational emotion analysis}\label{computational-emotion-analysis}

These insights render emotions an important object of computational anaylsis.
Computational emotion analysis involves measuring expressed or perceived emotional states through discrete emotion categories (e.g., anger, fear, joy, cf. Widmann and Wich 2022) or continuous dimensions, such as valence or arousal (e.g., Dietrich, Hayes, and O'brien 2019; Cochrane et al. 2022; Rittmann, Ringwald, and Nyhuis 2025).

Much research has relied on textual data to study political elites' emotional expression using methods such as sentiment dictionaries, word embeddings, or transformer models (Rheault et al. 2016; Gennaro and Ash 2022; Proksch et al. 2019; Osnabrügge, Hobolt, and Rodon 2021; Widmann and Wich 2022).
However, researchers increasingly turn to audio and visual modalities because of the various channels through wich emotions manifest in communication.
Audio-based studies extract acoustic features -- especially pitch -- to estimate speakers' emotional states (Dietrich, Enos, and Sen 2019; Dietrich, Hayes, and O'brien 2019; Rask and Hjorth 2025).
Image-based approaches detect facial expressions and body posture (Joo, Bucy, and Seidel 2019; Boussalis and Coan 2021; Boussalis et al. 2021; Torres 2024; Rittmann 2024).
Video-based research integrates these channels, analyzing facial and vocal expressions in motion (Rheault and Borwein 2019; Tarr, Hwang, and Imai 2022).

\subsection{Measuring arousal in political speeches}\label{measuring-arousal-in-political-speeches}

Arousal is of particular interest because it reflects the intensity with which emotions are expressed (Schlosberg 1954; Russell 1980; cf. Cochrane et al. 2022).
Unlike valence, which can often be reliably inferred from text alone (Rheault et al. 2016; Atteveldt, Velden, and Boukes 2021), arousal manifests in nonverbal cues such as tone, pitch, facial expression, and body movement (Dietrich, Hayes, and O'brien 2019; Rask and Hjorth 2025; Cochrane et al. 2022; Rittmann, Ringwald, and Nyhuis 2025; Rittmann 2024).
This implies that valid measurement of arousal requires multimodal approaches that integrate verbal, vocal, and visual channels.

Evidence by Cochrane et al. (2022) substantiates this argument.
In their study, Cochrane et al. (2022) distributed short segments of speeches held during \emph{Question Time} debates in the Canadian \emph{House of Commons} to two groups of coders for emotional valence and arousal rating.
One group rated speeches based on transcripts, the other based on video recordings, and both achieved moderate to high inter-rater reliability (see Table~\ref{tbl-cochrane_icc3k}).
However, their data shows that while human coders judge sentiment similarly when coding transcripts and videos, their text and video-based arousal ratings essentially measure different constructs.\footnote{The cross-rater average sentiment ratings based on transcripts and video recordings are strongly positively correlated (Pearson's \(r=0.711\), \(\text{CI}=[0.670, 0.748]\), \(N=627\)), whereas average arousal ratings are only very weakly correlated between annotation modalities (Pearson's \(r=0.119\), \(\text{CI}=[0.042, 0.196]\), \(N=627\); see Figure~\ref{fig-cochrane_crossmod_correlation}).
}
Given how much audio-visual cues contribute to arousal perception, this finding underscores the limitations of text-based analysis for capturing emotional intensity in political speech.

In addition to validity challenges, measuring arousal at scale is difficult because collecting human annotations is costly.
And wile automated approaches offer greater scalability (Rheault and Borwein 2019; Bagdon et al. 2024; Rittmann, Ringwald, and Nyhuis 2025), they face their own limiations.
Uni-modal methods, like vocal pitch analysis (Dietrich, Hayes, and O'brien 2019; Rask and Hjorth 2025) or facial expression detection (Boussalis and Coan 2021; Rittmann 2024), capture only a subset of relevant cues, and current multimodal approaches rely on supervised learning that requires domain-specific training data (Rheault and Borwein 2019; Tarr, Hwang, and Imai 2022; Rittmann, Ringwald, and Nyhuis 2025; Arnold and Küpfer 2025; but see Lüken et al. 2024).

\subsection{The promise of multimodal LLMs}\label{the-promise-of-multimodal-llms}

The emergence of multimodal large language models (mLLMs) enables a methodological shift towards \emph{in-context learning} for computational data annotation (Brown et al. 2020).
A model receives an annotation task instruction in the form of a natural-language prompt and generates a textual responses that is treated as an annotation, without task-specific finetuning.
This advantage has already driven wide adoption of in-context learning in text-based applications (e.g., Gilardi, Alizadeh, and Kubli 2023; O'Hagan and Schein 2023; Mens and Gallego 2024; Wu et al. 2024; Benoit et al. 2025; Licht et al. 2025; Ornstein, Blasingame, and Truscott 2025).

In-context learning with mLLMs opens new possibilities for computational emotion analysis but also comes with limitations that researchers must weigh.
Specifying theoretically grounded definitions directly in natural-language prompts and using instruction-following models to generate measurements appears promising in this application because research has shown that LLMs possess rich and nuanced internal representations of emotion constructs (Lee et al. 2025; Zhang and Zhong 2025).
However, relying on mLLMs for emotion analysis can also come at the cost of interpretability because it can be difficult to trace empirically how and which input features drive an (m)LLM to generate certain responses.
This barrier to interpretability is a limitation that in-context learning with mLLMs shares with other deep learning methods (Tarr, Hwang, and Imai 2022; Rittmann, Ringwald, and Nyhuis 2025; Arnold and Küpfer 2025; Lüken et al. 2024).

Yet, unlike supervised multimodal models, in-context learning with mLLMs requires no domain-specific training data or specialized hardware and enables construct measurement through natural-language prompts.
It thus substantially lowers barriers to multimodal research, particularly when relying on API-served commercial models.
However, relying on mLLMs can also come at the cost of the tractability and replicability of research findings.
GPU computations can only be reproduced exactly under very constraining circumstances (e.g., greedy decoding and a fixed compute environment, see Section~\ref{sec-empirical_strategy-task}).
And models served through APIs can be updated or deprecated by vendors (Barrie, Palmer, and Spirling 2025).

In sum, in-context learning with mLLMs promises conceptually guided multimodal analysis at scale, but researchers must weigh whether its methodological advantages outweigh their risks and limitations.
The methodological framework described next enables researchers to empirically inform this judgment.

\section{Empirical Strategy}\label{sec-empirical_strategy}

This paper examines the in-context learning capabilities of mLLMs available as of early 2026 by evaluating how well video-based scoring of speakers' emotional arousal aligns with human coders' ratings.
I use zero- and few-shot in-context learning as described in Section~\ref{sec-mllm_incontext_learning} and compare generated arousal ratings to human coders' ratings.

\subsection{In-context learning with mLLMs}\label{sec-mllm_incontext_learning}

\begin{figure}[!th]
    \caption{%
        Illustration of video-based arousal rating through mLLM in-context learning with token-probability weighted scoring.%
        \emph{Note:} see Table~\ref{tab:notation} for an overview of the notation.
    }
    \label{fig:diagram}
    \begin{minipage}{0.4\textwidth}
        \centering
        \includegraphics[width=\textwidth]{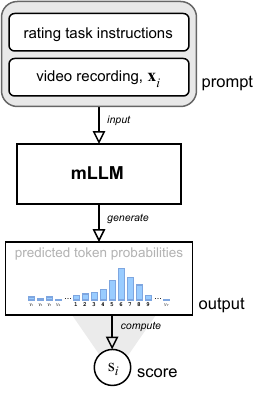}
    \end{minipage}%
    \begin{minipage}{0.6\textwidth}
        \begin{promptbox}[title={Example task instructions}]
\begin{lstlisting}
Below you are presented with the video recording of a person speaking.

Your task is to rate the **intensity of the emotions expressed by the speaker** in the recording on a scale from 1 (low/weak) to 9 (high/strong).

...

Return your rating as an integer value in the range 1 to 9. Only return your rating and nothing else.
\end{lstlisting}
        \end{promptbox}
    \end{minipage}%
\end{figure}

Figure \ref{fig:diagram} illustrates in-context learning with mLLM for video-based scoring.
I task an mLLM to \emph{rate} the emotional intensity a speaker expresses in a video recording.
The task instruction defines the construct (emotional arousal), the rating scale (e.g., 1--9),\footnote{I instruct mLLMs to respond with integer numbers instead of allowing decimal numbers because (a) this parallels human raters' task and (b) there are typically no native (single-token) representations for decimal numbers in LLMs' vocabularies.} and the response format.
The instruction text and a video recording (\(\mathbf{x}_i\)) form the prompt \(\mathbf{c}_i\) inputted to the model.

The model generates its response, \(y_i\), by estimating token probabilities, \(\Pr(y_i=w \mid \mathbf{c}_i)\), conditional on the prompt context \(\mathbf{c}_i\).
Here, \(w\) represents any individual token from the model's vocabulary \(\mathcal{V}\).\footnote{Put simply, an (m)LLM's vocabulary is the set of text pieces (tokens) it knows and can output (e.g., words, subwords, numbers, punctuation, symbols, etc.).}
Importantly, the model's vocabulary \(\mathcal{V}\) includes \(\mathcal{S}\), the set of tokens corresponding to the integer numbers that represent the available scale point options (e.g., \(\mathcal{S} = \{\texttt{1}, \ldots, \texttt{9}\}\)).
I use the model's token probabilities of scale point option tokens to compute a \textbf{probability-weighted rating score} (Licht et al. 2025; cf. Wang, Zhang, and Choi 2025; Ornstein, Blasingame, and Truscott 2025):
\begin{equation}\label{eqn:prob_weighting}
    s_i = \frac{1}{p_s}\sum_{s \in \mathcal{S}} \Pr(y_i=s \mid \mathbf{c}_i) \cdot \operatorname{n}(s),
\end{equation}
where \(p_s=\Pr(y \in \mathcal{S} \mid \mathbf{c}_i)\) is the total probability mass assigned to tokens in the scale \(\mathcal{S}\) (normalized to ensure valid probabilities), and \(\operatorname{n}()\) maps tokens to their integer representations.\footnote{This approach avoids the bunching Licht et al. (2025) observe in LLM's direct scalar responses in rating tasks that also shows in the mLLM responses generated in my analyses.}
The resulting measurement, \(s_i\), is treated as the mLLM's emotional arousal annotation of video \(\mathbf{x}_i\).

\subsection{Models}\label{sec-empirical_strategy-models}

Video-based in-context learning requires mLLMs that can process temporally aligned audio-visual input sequences.
I evaluate both closed- and open-weights mLLMs meeting this requirement.

For closed-weights models, I focus on Gemini 2.5 Flash and Gemini 3 Flash Preview -- the two most recent models in Google's Gemini Flash series as of January 2026, advertised as capable of ``video understanding'' (Google 2025c). Both allow controlling the degree of reasoning (``thinking'') during inference,\footnote{Gemini 3 Flash Preview provides the thinking modes ``minimal'' (`no thinking'), ``low'', ``medium'' and ``high'' (`dynamic' thinking).
  Gemini 2.5 Flash allows configuring a thinking budget ranging from 0 to 24,576 tokens, where a value of 0 corresponds to no reasoning and -1 is the (default) dynamic reasoning mode equivalent to ``high'' in Gemini 3.
  See Google (2025b).
  \label{fn:gemini_reasoning_control}} enabling evaluation under no-reasoning and dynamic-reasoning modes.

While closed-source models often lead in performance, they lack transparency and limit reproducibility (Barrie, Palmer, and Spirling 2025).
As a point in case, Gemini's documentation provides little detail about whether videos are processed as aligned audio-visual sequences rather than textual summaries.

Accordingly, I also evaluate open-weights mLLMs, focusing on the Qwen Omni (Xu, Guo, He, et al. 2025; Xu, Guo, Hu, et al. 2025) and TowerVideo (Viveiros et al. 2025) families.\footnote{These models are each available in different sizes and variants:
  Qwen 2.5 Omni in 7B and 3B parameter versions,
  Qwen 3 Omni in \emph{Thining} and (non-reasoning) \emph{Instruct} variant,
  and TowerVideo in 9B and 2B parameter versions}
Qwen 2.5 Omni (released on March 26, 2025) introduced architectural innovations like cross-modal self-attention, time-aligned multimodal RoPE, and time-interleaving (Xu, Guo, He, et al. 2025), while its successor Qwen 3 Omni (released on September 15, 2025) built upon these foundations with Mixture-of-Experts designs that enable dynamic reasoning similar to Gemini Flash models.
TowerVideo models (released on October 15, 2025) represents a separate architectural approach optimized for video understanding.

All examined mLLMs represent first-generation to early second-generation mLLMs specifically designed for video-based tasks.
This distinguishes them from earlier audio- and image-only models and video-text-to-text mLLMs that process videos without audio tracks (e.g., LLaVA, Janus, Video-R1, cf. Kumar 2025).

\subsection{Data}\label{sec-empirical_strategy-data}

I use two complementary datasets with human-labeled video recordings for mLLM evaluation.
First, the \textbf{RAVDESS} dataset (Livingstone and Russo 2018) contains 1,248 short videos of 24 speech actors (12 male, 12 female).
Livingstone and Russo (2018) instructed actors to perform one of two statements\footnote{`Dogs' (``Dogs are sitting by the door'')
  and
  `Kids' (``Kids are talking by the door'').
  The RAVDESS dataset also contains a \emph{song} module, but I only use the \emph{speech} module here.} in a ``Neutral'' state or one of seven emotions\footnote{Angry, Disgust, Fearful, Happy, Neutral, Sad, and Surprise.
} at normal or strong intensity.\footnote{Livingstone and Russo (2018) recorded two trials per actor, statement, emotion category, and intensity level.}
Each recording was rated by ten annotators on a 1--5 intensity scale and
I use their cross-coder averages of these rating for evaluation, as they exhibit high measurement reliability.\footnote{Livingstone and Russo (2018) report a an intra-class correlation (ICC) coefficient estimate of 0.74 (cf. Cicchetti 1994).}

Second, I use the \textbf{Parliamentary Speech Emotions} dataset contributed by Cochrane et al. (2022), specifically, the video-based arousal ratings of English speech segments in this dataset.
To generate benchmark measurements, I averaged each video coder's repeated arousal ratings\footnote{Cochrane et al. (2022) let coders rate some segments multiple times at different points to compute intra-coder reliability.} and then averaged these at the speech segment level for all 595 speeches.\footnote{I attempted to download video recordings of all 635 English-language segments but succeeded for only 595.}
This yields highly reliable measurements according to intra-class correlation (ICC) estimates (see Table~\ref{tbl-cochrane_icc3k}).

The Parliamentary Speech Emotions dataset ideally complements RAVDESS.
Instead of scripted emotions in controlled conditions, it captures politicians' speeches in live parliamentary debates where the focal speaker may not face the camera and ambient sounds may confound arousal indicators (see Section~\ref{sec-cochrane_data}).
Accordingly, it presents a more challenging but practically realistic test case.

I have split both datasets into train, validation, and test splits blocking by speaker\footnote{Blocking by speaker means that all the recordings of a given speaker are in the same split (train, validation, or test) and ensures out-of-sample evaluation also in few-shot inference.\textless!--even when using few-shot exemplars.}.
I use the test splits for in-context learning inference and evaluation, the train splits for exemplars sampling in few-shot in-context learning (see details below), and set aside the
remaining recordings for future studies examining, for example, fine-tuning approaches.
This yields 624 test and 312 train examples in the RAVDESS data
and
314 test and 145 train examples in the Cochrane et al. (2022) data.
Figure~\ref{fig-arousal_histograms_testsets} depicts the distribution of arousal measurements in the test splits of both datasets.

\subsection{Task}\label{sec-empirical_strategy-task}

I follow the procedure described in Section~\ref{sec-mllm_incontext_learning} to generate emotional arousal ratings for video recordings using zero- and few-shot in-context learning with mLLMs.\footnote{A caveate of the closed-weights Gemini model is that I can only access the log probabilities of the 19 most likely tokens at each generation step.
  In the case that any of the scale point option tokens in \(\mathcal{S}\) is not in this set, I replace the missing value with 0.0 to allow computing \ref{eqn:prob_weighting}.}
The prompt templates I use (see Prompt \ref{prompt:ravdess_arousal} and \ref{prompt:cochrane_arousal}) are aligned with the original coding instructions and follow best practices in prompt engineering.\footnote{I do not report results for different prompts, however, for computational reasons.}

In the RAVDESS data, I instruct mLLMs to rate each video on a 1--5 scale in terms of emotion intensity, paralleling the original human annotation task.
In the Parliamentary Speech Emotions data, I use a 1--9 scale, deviating from the 0--10 scale originally used because most models' tokenizers have no token for ``10'' and using the 1-9 scale avoids resulting computational complications.
Accordingly, I rescaled the average human rating measurements to the 1--9 range for evaluation.

In the zero-shot setting, only the task instruction and the focal video are provided.
Additionally, I run experiments with three and five exemplars sampled from the respective datasets' train splits.\footnote{To demonstrate the given model the range of possible input--score pairs, I have selected so-called ``anchor'' exemplars placed on about equidistant locations along the empirical range spanned by train set examples' scores.
  For instance, in 3-shot inference, this means that exemplars feature one high-, one moderate, and one low-scoring exemplar.}
Few-shot exemplars are integrated as turns of user input (video plus short task summary) and assistant response in the conversation history, before the to-be-rated video.
As the assistant's response, I use the cross-coder average arousal/intensity score rounded to integers to match the response format.

\subsubsection{Reproducibility}\label{reproducibility}

I run all examined open-weights mLLMs in 4-bit quantization on local hardware (see Section~\ref{sec-gpu_configs}) with greedy decoding using the \texttt{transformers} (Wolf et al. 2020) library, making generation deterministic given a fixed compute environment and batch (but see Section~\ref{sec-reproducibility_impact}).
Inference with Gemini Flash models was run through Google's Vertex AI API with temperature set to 0 in no-reasoning mode to approximate greedy decoding.
A more detailed discussion of reproducibility measures can be found in Section~\ref{sec-reproducibility} of the Supplemental Materials.

\subsection{Evaluation}\label{sec-empirical_strategy-evaluation}

The goal of my analyses is to understand how well video-based in-context learning with an mLLM measures a speaker's emotional arousal in speech recordings.
To provide a comprehensive assessment by examining overall reliability and systematic bias.

\subsubsection{Reliability}\label{reliability}

I evaluate mLLMs' scoring reliability using three quantitative evaluation metrics: the Pearson's \(r\) correlation coefficient, Spearman's \(\rho\) rank correlation coefficient, and the \emph{Root Mean Squared Error} (RMSE) between reference and LLM scores.
Pearson's \(r\) allows comparisons to ICC-based human inter-rater reliability estimates.
Spearman's \(\rho\) measures how similarly an mLLM ranks speakers' arousal relative to average human ratings.
And the RMSE measures average scoring error magnitude, reported as RMSE\(_{0–1}\) on the normalized 0--1 scale to facilitate comparison across datasets.\footnote{I have normalized reference and mLLM scores to the interval 0--1 using min-max normalization before computing the RMSE:
  \(\text{minmax}(x, l, h) = (x, l) / (h - l)\), where \(l\) and \(h\) are a scale's theoretical extrema (e.g., 1 and 5 in the RAVDESS data).}

However, evaluating mLLMs against average human ratings introduces \emph{attenuation} (Schmidt and Hunter 1999).
Because we do not observe the ``true'' level of emotional arousal in a speech and human raters' average ratings exhibit measurement error,\footnote{For example, the ICC1k (ICC3k) estimates for video-based emotion intensity ratings in the RAVDESS (Cochrane et al.) data of 0.74 (0.83) imply that 26\% (17\%) of the variance in the averaged ratings used as reference scores are due to annotation noise.} even a model that perfectly measures speeches' unobserved arousal levels would not correlate perfectly with these reference scores.
I therefore account for attenuation when discussing evaluation results.\footnote{Following Schmidt and Hunter (1999), I assume \(r_\text{obs} = r_\text{true} \sqrt{ \rho_\text{humans} \rho_\text{model} }\).
  Given fixed inputs and greedy decoding, \(\rho_\text{model} \approx 1\) and \(r_\text{true} = r_\text{obs} / \sqrt{ \rho_\text{humans} }\).
  Note that this approach is conservative, however, as setting \(\rho_\text{model} < 1\) would \emph{increase} \(r_\text{true}\).}

\subsubsection{Bias analysis}\label{sec-empirical_strategy-bias_analysis}

The quantitative evaluation metrics I report summarize an mLLM's overall scoring performance relative to average human ratings.
However, they do not reveal whether its arousal scores exhibit systematic biases across demographic groups.
In particular, I focus demographic groups defined by \emph{gender} here because it is observed in both the RAVDESS and the Parliamentary Speech Emotions datasets and it tends to manifest in observable audio-visual cues that might influence mLLMs inference.

To assess gender bias in the emotional arousal scores an mLLM generates, I rely on a decomposition of its empirical prediction errors into systematic bias and random noise components to compare the magnitude and direction of its systematic bias in videos by male and female speakers.
The error decomposition relies on the identity:
\begin{equation}\label{eqn:decomposition_identity}
\text{MSE} = \underbrace{(\text{MPE})^2}_{\text{bias component}} + \underbrace{\text{Var}(\mathbf{s} - \mathbf{y})}_{\text{noise component}}
\end{equation}
where
\(\mathbf{s}\) are the mLLM's arousal scores of videos;
\(\mathbf{y}\) their human rating-based reference scores;
MSE is the \emph{Mean Squared Error}, \(\frac{1}{n} \sum_{i=1}^n (s_i - y_i)^2\);
MPE is the \emph{Mean Prediction Error}, \(\frac{1}{n} \sum_{i=1}^n (s_i - y_i)\);
and \(\text{Var}(\cdot)\) denotes the variance operator.\footnote{This decomposition follows because for any random variable \(X\), \(E[X^2] = \text{Var}(X) + (E[X])^2\).
}
The MPE measures \emph{bias} because it is a directional indicator of systematic scoring error: positive values indicate a tendency to over-predict relative to average human ratings, negative values indicate under-prediction.
Crucially, focusing on the MPE rather than RMSE-type metrics isolates systematic scoring error from random noise, ensuring that any observed differences across demographic groups cannot be attributed to variance differences alone.

I then compute gender-specific MPEs (MPE\textsubscript{male} and MPE\textsubscript{female}) separately for videos of male and female speakers.
Subtracting an mLLM's MPE\textsubscript{male} from its MPE\textsubscript{female} allows me to assess whether its scores exhibit a \emph{net scoring bias} attributable to speaker gender: a positive (negative) difference indicates that the net bias lets female speakers appear more (less) aroused than male speakers compared to average human ratings.
To assess whether the observed differences in MPEs across speaker gender groups is statistically significant, I rely on bootstrapped resampling to compute and subtract multiple group-specific MPEs, and then perform a two-sample \(t\)-test comparing the distributions of MPE difference estimates.

\section{Results}\label{results}

\subsection{Results in the RAVDESS data}\label{sec-ravdess}

\begin{figure}[!th]

\centering{

\pandocbounded{\includegraphics[keepaspectratio]{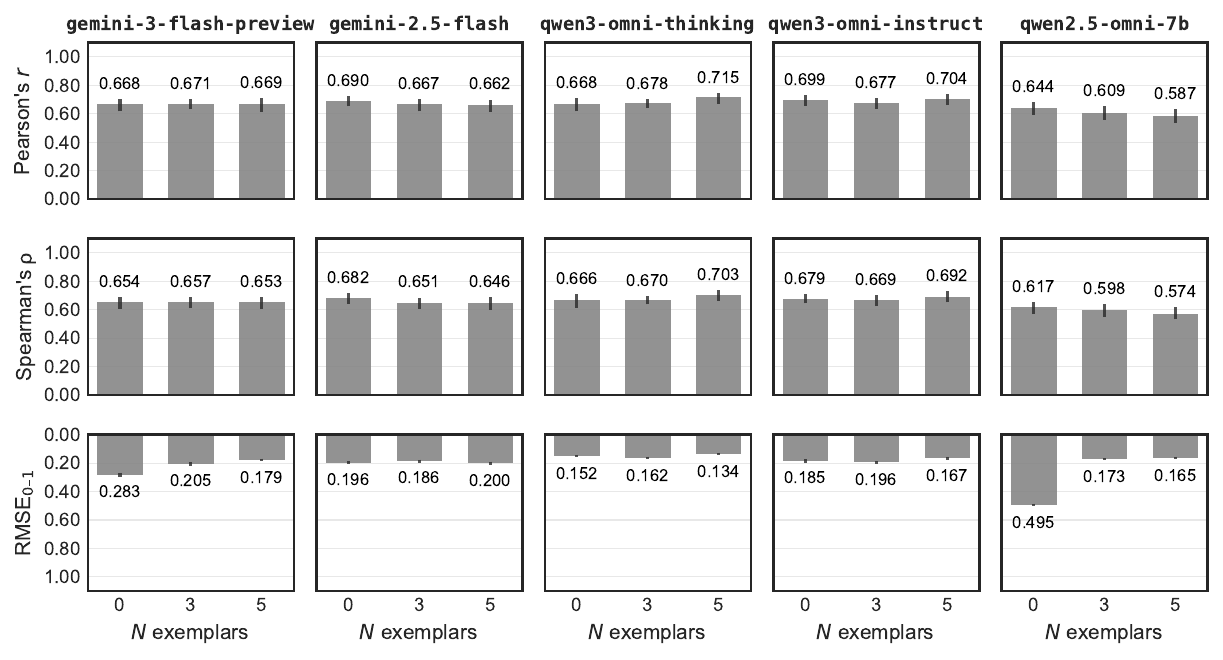}}

}

\caption{\label{fig-ravdess_scoring_res}Video-based emotion intensity scoring performance of mLLMs in the RAVDESS data test split. Performance, shown on y-axes, measured in terms of correlation (ρ, \(r\)) and the RMSE on the normalized 0--1 scale (RMSE\(_{0–1}\)) is computed against average human ratings; higher (lower) values in ρ and \(r\) (RMSE) indicate better performance. Bar height depicts bootstrapped averages (N=120); vertical black show 90\% confidence intervals. x-axis values indicate the number of few-shot ICL examples (0, 3, or 5).}

\end{figure}%

Figure~\ref{fig-ravdess_scoring_res} reports the performance of selected mLLMs in scoring emotion intensity in the RAVDESS data test split.\footnote{Results for all examined mLLMs, including TowerVideo models and Qwen 2.5 Omni 3B, are reported in Table~\ref{tbl-ravdess_scoring_res}.}
Note that in the case of reasoning models, I present results from inference with no or minimal thinking (see footnote \ref{fn:gemini_reasoning_control}), as this yielded better performance in this task and data (see Table~\ref{tbl-ravdess_comp_thinking_scoring_res}).
The open-weights Qwen 3 Omni Thinking (without reasoning) achieves the strongest results -- closely followed by its Instruct counterpart --, reaching a Pearson correlation of 0.715 and an RMSE\(_{0–1}\) of 0.134 in 5-shot inference.\footnote{To contextualize the RMSE\(_{0–1}\) estimates, using random guessing (sampling from a uniform distribution) as a baseline yields an RMSE\(_{0–1}\) of 0.408; using the average reference score values as prediction yields an RMSE\(_{0–1}\) of 0.145 in this data.}

Adjusted for attenuation, this corresponds to a correlation of 0.831, exceeding the estimated reliability of human coders' arousal ratings reported by Livingstone and Russo (2018).
Among the closed-source Gemini models, Gemini 3 Flash Preview performs best, achieving a correlation of 0.671 and RMSE\(_{0–1}\) of 0.205 in 3-shot inference with ``minimal'' reasoning (attenuation-adjusted correlation: 0.780), closely followed by Gemini 2.5 Flash.

\begin{figure}[!th]

\centering{

\pandocbounded{\includegraphics[keepaspectratio]{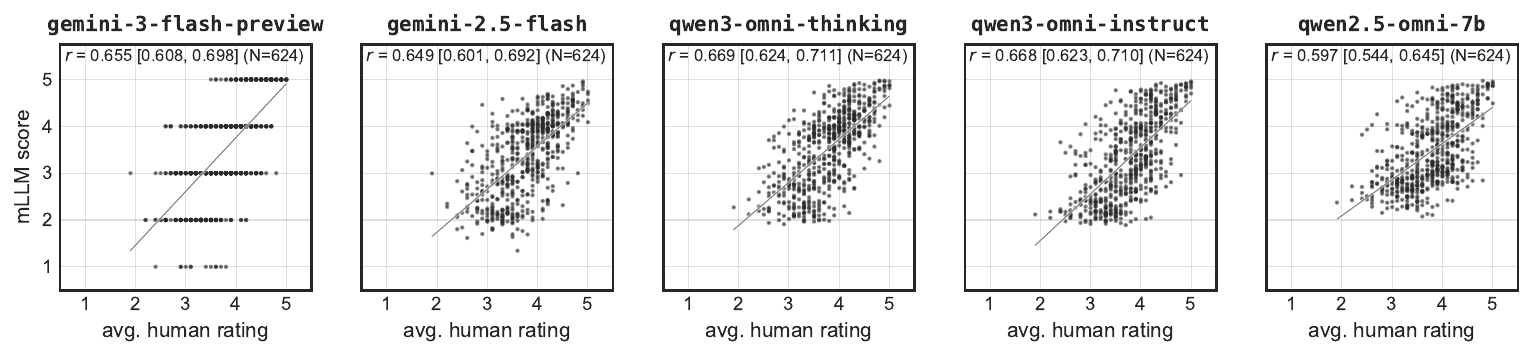}}

}

\caption{\label{fig-ravdess_scoring_scatter}Relationship between mLLM's video-based 3-shot emotion intensity scores and human coders' average ratings in the RAVDESS data test split. Correlation is pearson's \(r\). \emph{Note:} Values deviate from Figure~\ref{fig-ravdess_scoring_res}, which reports bootstrapped averages rather than point estimates.}

\end{figure}%

Figure~\ref{fig-ravdess_scoring_scatter} shows the relation between model and reference scores in 3-shot inference for these mLLMs.\footnote{Figure~\ref{fig-ravdess_scoring_scatter_all} reports scatter plots for all examined models.}
The scatter plots reveal distinct scoring patterns across models.
All mLLMs except Gemini 3 Flash Preview produce well-calibrated scores with continuous distributions that closely align with the reference scores.
Further, Figure~\ref{fig-ravdess_scoring_boxplots} shows that these models recover intensity differences between videos in the ``Normal'' and ``Strong'' conditions.\footnote{Both human and mLLM scores show: (i) low ratings for ``Neutral'' emotions, (ii) higher intensity ratings for ``Strong'' vs.~``Normal'' conditions, and (iii) variable emotion-specific intensity differences.}
In contrast, Gemini 3 Flash Preview' exhibits discrete scoring patterns, producing scores closely clustered around the integer rating scale option visually perceptible as horizontal bands in its scatter plot.
This is because we cannot switch off reasoning completely in this model version and reasoning tends to generated collapsed token probability distributions (see Figure~\ref{fig-ravdess_response_token_probability_distributions_thinking_models}).

\subsubsection{Gender bias analysis}\label{gender-bias-analysis}

Besides overall performance, a key question is whether mLLMs' scores are biased with respect to speakers' demographic characteristics.
Using the error decomposition approach described in Section~\ref{sec-empirical_strategy-evaluation}, I compare the systematic bias component in videos of male and female speakers.

\begin{table}

\caption{\label{tbl-ravdess_bias_test_all}Bias analysis of mLLMs' video-based emotion intensity scores in the RAVDESS data test split, overall and by speaker gender. MPE\(_{0–1}\) in videos of male and female speakers indicates whether a model over- or underestimates relative to human ratings. Test statistics (\(t\), \(p\)) and effect size (Cohen's \(d\)) report a two-sided \(t\)-test for the MPE difference between male vs.~female speakers across 120 bootstrapped samples. \emph{Note:} MPE values deviate from Table~\ref{tbl-ravdess_error_decomposition_all}, which reports bootstrapped averages rather than point estimates.}

\centering{

\centering%
\normalsize%
\begin{tabular}{rrrrrrr}
\toprule
 & MPE\textsubscript{male} & MPE\textsubscript{female} & $\Delta$ MPE & $t$-statistic & $p$-value & Cohen's $d$ \\
\midrule
qwen3-omni-thinking & -0.064 & -0.062 & 0.002 & -1.480 & 0.140 & 0.192 \\
gemini-3-flash-preview & -0.074 & -0.063 & 0.011 & -7.581 & 0.000 & 0.983 \\
qwen2.5-omni-7b & -0.107 & -0.058 & 0.049 & -41.772 & 0.000 & 5.415 \\
gemini-2.5-flash & -0.138 & -0.073 & 0.066 & -56.809 & 0.000 & 7.365 \\
qwen3-omni-instruct & -0.157 & -0.077 & 0.080 & -66.243 & 0.000 & 8.588 \\
\bottomrule
\end{tabular}

}

\end{table}%

Table~\ref{tbl-ravdess_bias_test_all} reports estimates of the systematic bias of mLLMs examined in 3-shot inference computed separately in male and female speakers' recordings in the RAVDESS test split.\footnote{Across mLLMs and speaker gender groups, the systematic bias component accounts for between 9.2\% and 52.8\% of total scoring error (see Table~\ref{tbl-ravdess_error_decomposition_all}); the wide range is primarily driven by variation in the degree of gender-differential bias across models.}
It shows that, on average, the examined mLLMs tend to underestimate emotional intensity relative to human ratings (see also Table~\ref{tbl-ravdess_error_decomposition_all}) and \emph{more} so for male speakers than for female speakers.
Despite exhibiting a downward bias for both groups, the stronger underestimation of male speakers results in a net-positive gendered intensity rating bias that makes female actors appear more aroused than male actors when compared to average human ratings.
Computing the difference between a model's MPE estimates in male and female actors' recordings and testing this difference for significance with two-sided \(t\)-tests computed from 120 bootstrapped samples, shows that all examined mLLMs except Qwen 3 Omni Thinking exhibit varying but significant degrees of differential bias by speaker gender.\footnote{Linear regression modeling of mLLM score residuals (relative to human-rating reference scores) lead to substantively similar conclusions, although the magnitude of standard errors and hence statistical significance for the speaker gender coefficient estimate varies depending on the modeling strategy (clustered standard errors or not, OLS or mixed-effects model).}
The MPE differences for these models range from a modest 0.011 on the 0--1 scale (Gemini 3 Flash Preview) up to a notable difference of 0.08 (Qwen 3 Omni Instruct, see Table~\ref{tbl-ravdess_error_decomposition_all}).
The error decomposition further shows that this gender-differential pattern is driven by differences in the \emph{systematic bias} component rather than random noise (see Table~\ref{tbl-ravdess_error_decomposition_all}).
For models exhibiting significant gender-differential bias, the share of scoring error attributable to systematic bias is consistently higher for male speakers than for female speakers -- ranging from 35.9\% to 52.8\% for male speakers compared to 12.1\% to 18.9\% for female speakers in the three most affected models.

Since my analyses use average human ratings as a reference, these findings should be assessed in light of potential gender bias in raters' annotations.
Leveraging the RAVDESS data's controlled laboratory design, I compare human raters' intensity ratings of male and female actors via linear regression controlling for all observed experimental factors (emotion, intensity stimuli, statement, and repetition conditions; see Section~\ref{sec-ravdess_gender_bias_human_ratings}), finding that female actors' intensity is rated \textasciitilde0.055 points higher on the 0--1 scale (see Table~\ref{tbl-rating_gender_bias_coefs}).
If we assume that this between-gender difference in average human ratings reflects female actors' truly more intense emotional display within experimental conditions (e.g., due to socialization and gendered norms of emotional display), a net-positive gendered intensity rating bias in an mLLM's scores \emph{introduces} gender bias in measurements.
If, instead, we assume that the between-gender intensity rating difference observed in average human ratings is due to raters' biased \emph{perception} of female actors' emotional display, a net-positive gendered intensity rating bias in mLLM scores not only introduces but adds to and compounds human raters' gender bias in measurements.

\subsubsection{Discussion}\label{discussion}

While the findings in Figure~\ref{fig-ravdess_scoring_res} and Figure~\ref{fig-ravdess_scoring_scatter} suggest cautious optimism regarding mLLMs' overall emotion intensity scoring performance, the indication of systematic -- though sometimes modest -- gender-differential bias raises concerns about their fairness in real-world applications.
I thus turn next to the examined mLLMs' perfomance in real-world parliamentary speech recordings.

\subsection{Results in the Parliamentary Speech Emotions data}\label{results-in-the-parliamentary-speech-emotions-data}

\begin{figure}[!th]

\centering{

\pandocbounded{\includegraphics[keepaspectratio]{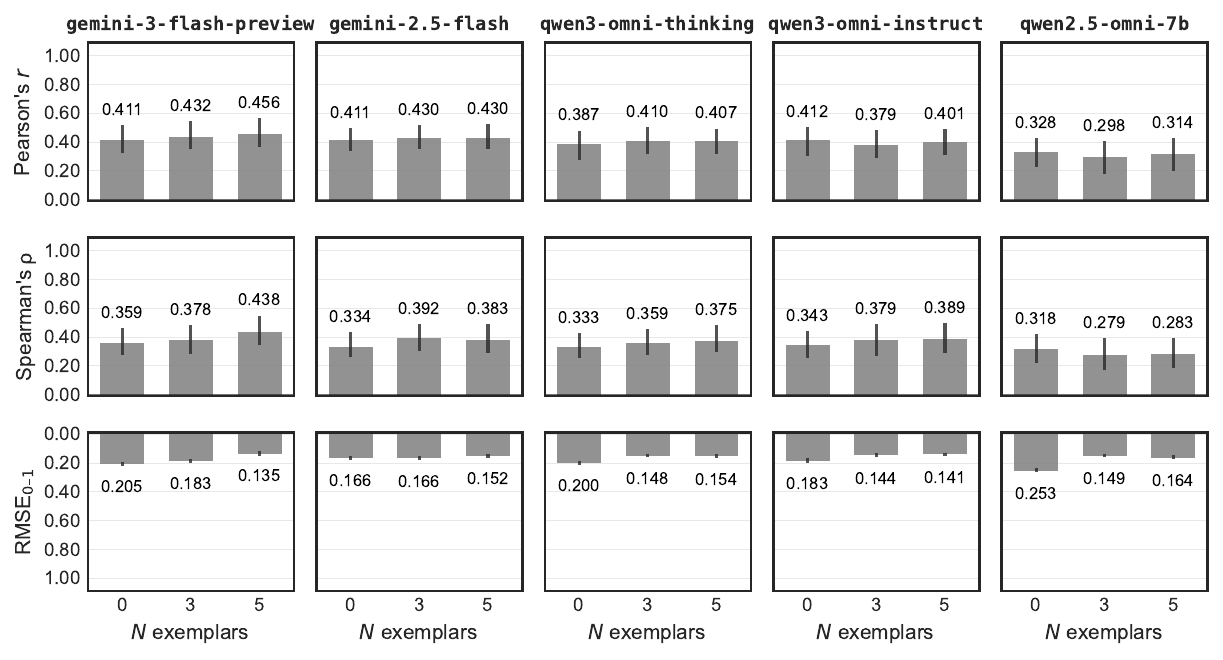}}

}

\caption{\label{fig-cochrane_scoring_res}Video-based emotional arousal scoring performance of mLLMs in the Parliamentary Speech Emotions data test split. Performance, shown on y-axes, measured in terms of correlation (ρ, \(r\)) and the RMSE on the normalized 0--1 scale (RMSE\(_{0–1}\)) is computed against average human ratings; higher (lower) values in ρ and \(r\) (RMSE) indicate better performance. Bar height depicts bootstrapped averages (N=120); vertical black show 90\% confidence intervals. x-axis values indicate the number of few-shot ICL examples (0, 3, or 5).}

\end{figure}%

Figure~\ref{fig-cochrane_scoring_res} reports selected mLLMs' video-based arousal scoring performance against human coding-based reference scores in the Parliamentary Speech Emotions data (Cochrane et al. 2022), with thinking mode disabled for reasoning mLLMs except Gemini 3 Flash Preview (see Table~\ref{tbl-cochrane_comp_thinking_scoring}).\footnote{Results for all examined mLLMs, including TowerVideo models and Qwen 2.5 Omni 3B, are reported in Table~\ref{tbl-cochrane_arousal_scoring_res}.}
All analyzed models perform comparatively poorly in the Parliamentary Speech Emotions data.
Correlations are low in absolute terms, even after adusting for attenuation.\footnote{Given ICC3k estimates for video-based arousal ratings and 0.831 (see Table~\ref{tbl-cochrane_icc3k}) the unattenuated Pearson's \(r\) metric estimate for the best-performing mLLM -- Gemini 3 Flash in 5-shot inference with reasoning enabled (\(r\)=0.456) -- is only 0.5.}
In particular, correlations are consistently lower than in the RAVDESS data independent of model type, size, or few-shot setting.
The comparatively low RMSE values are the exeption.\footnote{To contextualize the RMSE\(_{0–1}\) estimates, using random guessing (sampling from a uniform distribution) as a baseline yields an RMSE\(_{0–1}\) of 0.408; using the average reference score values as prediction yields an RMSE\(_{0–1}\) of 0.123 in this data.}
But Figure~\ref{fig-cochrane_arousal_scoring_scatter_plots_all} Figure~\ref{fig-cochrane_arousal_scoring_scatter_plots_all} shows they arise because mLLMs' scores cluster near the mean of the reference score distribution despite poor correlation.

\subsubsection{Gender bias analysis}\label{gender-bias-analysis-1}

\begin{table}

\caption{\label{tbl-cochrane_bias_test_all}Bias analysis of mLLMs' video-based emotion intensity scores in the Parliamentary Speech Emotions data test split, overall and by speaker gender. MPE\(_{0–1}\) in videos of male and female speakers indicates whether a model over- or underestimates relative to human ratings. Test statistics (\(t\), \(p\)) and effect size (Cohen's \(d\)) report a two-sided \(t\)-test for the MPE difference between male vs.~female speakers across 120 bootstrapped samples. \emph{Note:} MPE values deviate from Table~\ref{tbl-cochrane_error_decomposition_all}, which reports bootstrapped averages rather than point estimates.}

\centering{

\centering%
\normalsize%
\begin{tabular}{rrrrrrr}
\toprule
 & MPE\textsubscript{male} & MPE\textsubscript{female} & $\Delta$ MPE & $t$-statistic & $p$-value & Cohen's $d$ \\
model &  &  &  &  &  &  \\
\midrule
gemini-3-flash-preview & -0.041 & -0.006 & 0.034 & -18.584 & 0.000 & 2.409 \\
gemini-2.5-flash & -0.061 & -0.030 & 0.031 & -19.721 & 0.000 & 2.557 \\
qwen3-omni-thinking & 0.013 & 0.041 & 0.028 & -18.733 & 0.000 & 2.429 \\
qwen3-omni-instruct & -0.028 & -0.003 & 0.024 & -16.835 & 0.000 & 2.182 \\
qwen2.5-omni-7b & 0.031 & -0.007 & -0.039 & 25.289 & 0.000 & -3.278 \\
\bottomrule
\end{tabular}

}

\end{table}%

Further, I find that mLLMs' arousal scores in the Parliamentary Speech Emotions data are differentially biased by speaker gender (see Table~\ref{tbl-cochrane_bias_test_all}).
Similar to the results in the RAVDESS data, most of the examined mLLMs tend to systematically underestimate speakers' arousal in 3-shot inference relative to average human ratings -- and more so for male speakers than for female speakers (see Table~\ref{tbl-cochrane_bias_test_all}).\footnote{Because the parliamentary speeches recorded in the Parliamentary Speech Emotions data were not created under experimentally controlled conditions, I cannot empirically test for systematic human rating differences due to speakers' gender in this data.}
For example, Gemini 2.5 Flash's overall MPE is -0.049 on the 0--1 scale (8.8\% of total error; see Table~\ref{tbl-cochrane_error_decomposition_all}), with gender-specific MPEs of -0.061 (male) and -0.030 (female) -- a difference of 0.031 that is statistically significant (\(t\) = -19.72, \(p\) \textless{} 0.000).
As in the RAVDESS analyses, the error decomposition indicates this pattern is driven by the systematic bias component rather than random noise (see Table~\ref{tbl-cochrane_error_decomposition_all}).
The net effect of this pattern is that female speakers appear more aroused than male speakers when relying on the mLLM's ratings rather than human coders' average ratings.

The exceptions to this pattern are Qwen 3 Omni Thinking and Qwen 2.5 Omni (7b).
Qwen 3 Omni Thinking \emph{over}-estimates arousal for both speaker genders in 3-shot inference with dynamic thinking, but more so for female speakers, so the net direction of the gender gap is the same as in the other models -- female speakers appear more aroused than male speakers relative to average human ratings.
Qwen 2.5 Omni (7b), in contrast, overestimates only male speakers' arousal, and this results in \emph{net} bias in the opposite direction: male speakers appear more aroused than female speakers compared to average human ratings.

\subsubsection{Discussion and summary of additional analyses}\label{discussion-and-summary-of-additional-analyses}

The Parliamentary Speech Emotions results offer two sobering conclusions.
The examined mLLMs' arousal scores align at best moderately with average human ratings in real-world data.
And their scores exhibit gender-differential bias -- consistent with my findings in the RAVDESS data.

Additional analyses reported in the Supplemental Materials add nuance to these conclusions.
First, Section~\ref{sec-apx_cochrane_explanation_1}, explores one plausible reason for mLLM's comparatively poor overall performance in real-world speech recordings: that background movements and noise might hamper mLLMs' ability to accurately score emotional arousal from video recordings.
I scrutinize this explanation by mitigated the influence of these potential nuisances through image masking and audio denoising.
However, the results show that this does \emph{not} systematically improve the examined mLLMs' scoring performance, suggesting that data quality issues do \emph{not} (fully) account for the examined mLLMs' poor overall performance.

Second, Section~\ref{sec-apx_cochrane_sentiment_results} examines whether mLLMs' poor scoring performance in real-world parliamentary speech videos is specific to arousal measurement by replicating the experiments reported above for sentiment scoring.
This shows that while the examined mLLMs' video-based sentiment ratings align better with human scores than their arousal ratings (see Table~\ref{tbl-cochrane_sentiment_scoring_res}), video-based performance remains notably lower compared to the performance in text-based inference (holding constant model and few-shot setup).
This suggests systematic challenges in video-based inference persist beyond arousal scoring.

Third, because measuring a quantity of interest is often only the first step in applied research, Section~\ref{sec-apx_cochrane_downstream_application} presents an application example that examines the average difference in government and opposition speakers' arousal based on the Parliamentary Speech Emotions test split.
This analysis demonstrates that despite poor calibration, mLLMs' arousal scores tend to replicate a the finding that opposition speakers are more aroused than government speakers obtained in a regression analysis of average human ratings, especially when adopting design-based supervised learning (Egami et al. 2024) to correct for measurement error in the mLLM-based arousal scores.

\section{Conclusion}\label{conclusion}

This study provides the first systematic assessment of mLLMs' ability to measure emotional arousal in political speech videos through in-context learning, auditing the first generation of models capable of processing temporally aligned audio-visual inputs as of early 2026 -- including Gemini 2.5/3 Flash, Qwen 2.5/3 Omni, and TowerVideo.

Using two complementary datasets, I show that the examined mLLMs perform well in scoring arousal under controlled laboratory conditions but fail in real-world political debates, where even the best-performing mLLM achieves a Pearson's \(r\) of only 0.456 with human ratings.
Moreover, in each dataset, all but one model exhibits systematic gender-differential bias, underestimating arousal more for male speakers.
And even in sentiment scoring, where human annotations are highly reliable across modalities (Cochrane et al. 2022), mLLMs perform worse in video-based than in text-based inference.

For applied researchers, these findings warrant three specific precautions:
First, do not assume text-based LLM performance generalizes to video tasks.
Second, always validate mLLMs against human coding in your specific application context before deployment.
Third, be especially cautious when analyzing cases with demographic variation: most examined mLLMs systematically underestimate arousal more for male than for female speakers, and this bias may compound rather than merely reflect biases present in human annotations.

Importantly, these findings represent a temporal snapshot of mLLM capabilities as of early 2026.
As mLLM development progresses, future models may exhibit different performance characteristics.
This underscores why establishing a baseline now matters: future studies can use my framework to identify whether next-generation mLLMs close the lab-to-field gap documented here.

In particular, this study raises several questions for future research.
First, what factors hamper mLLMs' video-based emotion measurement in real-world recordings?
My evidence suggests signal-to-noise improvements alone do not suffice, but specific acoustic (background noise, overlapping speech) and visual properties (camera angle, speaker movement) deserve further investigation.
Second, do these findings extend beyond scalar rating tasks?
If mLLMs can classify but not scale emotions, this would have major implications for research design.
For example, future research could examine discrete emotion detection or emotion trajectories.
Third, short-form political content (TikTok, Instagram reels, YouTube shorts) poses distinct challenges through rapid editing and stylized affect.
As political communication increasingly migrates to these platforms, understanding mLLM performance in these formats will be crucial.

\bigskip

\section*{Acknowledgments}

The author made use of OpenAI's GPT-5 and Anthropic's Claude Sonnet 4.5 in agentic mode (accessed via GitHub Copilot plugin in VS Code, September 2025 -- January 2026) to support wording edits, organization, and code suggestions during data collection and visualization of analysis results.
The tool was used without modification, and all outputs were reviewed and verified by the author, who remains solely responsible for the scientific content and conclusions.

\bigskip

\section{References}\label{references}

\begingroup
\setstretch{1.15}

\phantomsection\label{refs}

\endgroup

\clearpage
\appendix
\setcounter{page}{1}
\begin{center}
\Huge \scshape Supplemental Materials \par
\LARGE \bfseries Computational emotion analysis with multimodal LLMs \par
\end{center}
\vfill
\renewcommand{\contentsname}{Contents}
\startcontents[appendix]
\printcontents[appendix]{}{1}{\setcounter{tocdepth}{3}}
\vfill
\clearpage

\addtocounter{secnumdepth}{1}
\counterwithin{figure}{section}
\counterwithin{table}{section}
\renewcommand\thefigure{\thesection\arabic{figure}}
\renewcommand\thetable{\thesection\arabic{table}}

\section{Notation}\label{notation}

\begin{table}[h]
\centering
\small
\caption{Notation for mLLM-based arousal scoring}\label{tab:notation}
\begin{tabular}{ll}
\toprule
Symbol & Definition \\
\midrule
$\mathbf{x}_i$ & video recording $i$ \\
$\mathbf{c}_i$ & full prompt context (instructions + video encoding) \\
$\mathcal{V}$ & model's complete vocabulary \\
$\mathcal{S} \subset \mathcal{V}$ & subset of tokens representing scale points \\
$w$ & Individual token from $\mathcal{V}$ \\
$s$ & scale point option token in $\mathcal{S}$ (scale point option) \\
$y_i$ & model's generated response for video $i$ \\
$\operatorname{n}()$ & Mapping from token strings to integers \\
$s_i$ & Final probability-weighted arousal score \\
\bottomrule
\end{tabular}
\end{table}

\clearpage

\section{Data}\label{data}

\subsection{Cochrane et al.'s Parliamentary Speech Emotions data}\label{sec-cochrane_data}

\begin{figure}[!th]

\begin{minipage}{0.50\linewidth}

\centering{

\pandocbounded{\includegraphics[keepaspectratio]{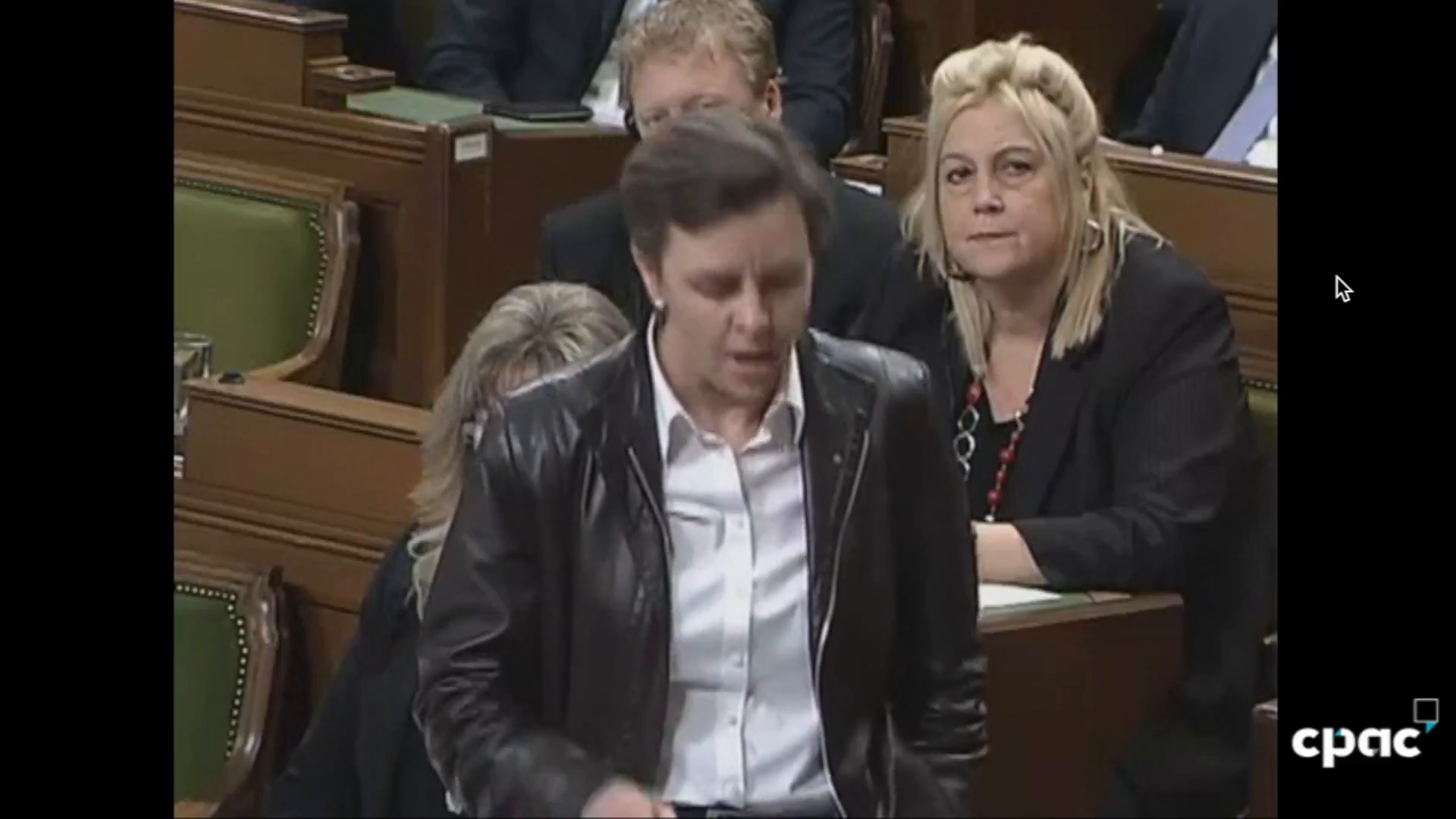}}

}

\subcaption{\label{fig-cochrane-example-frame-10}video frame at 00:00:10}

\end{minipage}%
\begin{minipage}{0.50\linewidth}

\centering{

\pandocbounded{\includegraphics[keepaspectratio]{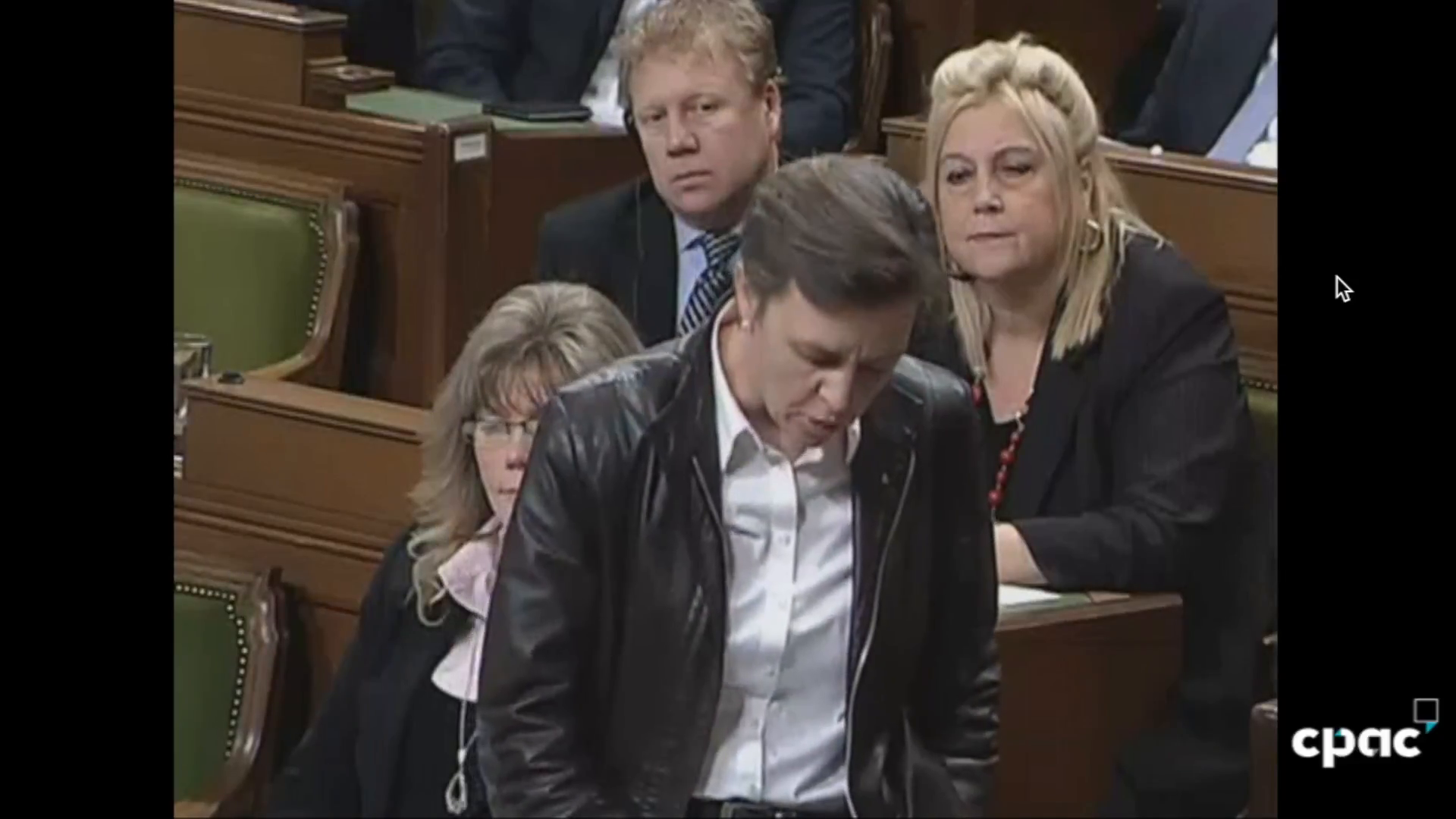}}

}

\subcaption{\label{fig-cochrane-example-frame-10}video frame at 00:00:40}

\end{minipage}%

\caption{\label{fig-cochrane-example-frames}Example of two frames of a video recording in the Cochrane et al.~(2022) data. (\href{https://www.youtube.com/watch?v=7wAJPC7Hw68}{original video link})}

\end{figure}

Figure~\ref{fig-cochrane-example-frames} shows two frames of a video taken from a video in the Cochrane et al. (2022) dataset that illustrated what realistic video recordings in this setting look like: the focal speaker may not face the camera, their posture may change throughout the recording, and there may be other people and objects in the background.
This is because in ``talking parliaments'' (Weber cited in Palonen 2018), speakers often speak where they sit so that the camera is often not frontally capturing them, which contrasts with video recordings from, for example, the German \emph{Bundestag} (Rittmann 2024).
Further, the audio track of the recording linked in the caption of Figure~\ref{fig-cochrane-example-frames} highlights that live recordings feature ambient sounds and background noises that may confound indicators of emotional arousal in the focal speaker's voice.

\subsubsection{Inter-coder reliability estimates}\label{inter-coder-reliability-estimates}

Table~\ref{tbl-cochrane_icc3k} reports the inter-rater agreement level in the Cochrane et al.~data by annotation modalities (text vs.~video) and emotion dimensions (sentiment vs.~arousal) in terms of the \emph{average fixed raters} intra-class correlation (ICC) coefficient (ICC3k, see McGraw and Wong 1996) for transcript and video-based sentiment and arousal ratings, respectively, computed from averages of coders' repeated ratings\footnote{Cochrane et al. (2022) let coders rate some segments multiple times at different points to compute intra-coder reliability.} per speech segment, dimension, and annotation modality, to estimate reliability.\footnote{I focus on the average of the given annotators' ratings -- in contrast to agreement between pairs of coders, like as Cochrane et al. (2022) -- because I am want to assess the reliability of the specific set of ratings produced by the fixed set of annototators (see McGraw and Wong 1996) and I use average of these annotators' ratings in my analyses.
  Generally, the \(k\)-coders \emph{average} ICC metrics ICC1k, ICC2k and ICC3k estimate the reliability of \(k\) annotators when working as a group.
  Put simply, they estimate how well the mean of \(k\) raters' scores would agree with the ``true'' score.
  In contrast, ICC1, ICC2 and ICC3 represent estimates of annotators individual reliabilities.}
The level of inter-rater reliability is overall moderate to high (Cicchetti 1994) across annotation modalities and emotion dimensions.
This conclusion holds when computing the ICC1k and ICC2k metrics instead (see Table~\ref{tbl-cochrane_other_iccs}),\footnote{Assessing ICC1k is adequate when the annotation panel is very unbalanced so that which annotator annotates which item varies a lot and, hence, one treats raters as a random factor.
  The ICC2k, in turn, is appropriate when assumes that raters are exchangeable one wants to know how reliable annotations would be with a new set of coders.}
ICC3k only in the subset of examples used for LLM scoring evaluation (Table~\ref{tbl-cochrane_icc3k_testset}),
or when using another multi-annotator reliability metric for interval/oridinal scaled annotations (Krippendorff's \(\alpha\), Table~\ref{tbl-cochrane_kalphas}).

However, comparing the ICC3k estimates between annotation modalities and emotion dimensions shows that inter-rater agreement is higher in text- than in video-based sentiment ratings whereas this pattern is reversed in arousal ratings.
This suggests that providing speeches' audio-visual information in the form of video recordings leads to slightly \emph{lower} reliability in \emph{sentiment} measurement but to \emph{higher} reliability in \emph{arousal} measurement.
This, in turn, suggests that in human annotation, text transcripts are better suited as annotation modality than videos in the case of sentiment measurement but \emph{less} suited in the case of arousal measurement.

\begin{table}

\caption{\label{tbl-cochrane_icc3k}Inter-coder agreement in text transcript- and video-based ratings of valence/sentiment and arousal/intensity according to the average fixed raters intra-class correlation (ICCk3) coefficient.}

\centering{

\centering%
\normalsize%
\begin{tabular}{lcc}
\toprule
 & Text ratings & Video ratings \\
\midrule
Sentiment & 0.937 [0.930, 0.940] & 0.832 [0.790, 0.870] \\
Arousal & 0.728 [0.700, 0.760] & 0.831 [0.790, 0.870] \\
\bottomrule
\end{tabular}

}

\end{table}%

\begin{table}

\caption{\label{tbl-cochrane_other_iccs}Inter-coder agreement in text transcript- and video-based ratings of valence/sentiment and arousal/intensity according to the average intra-class correlation (ICCk2 and ICCk2) coefficient.}

\centering{

\centering%
\normalsize%

\resizebox{\textwidth}{!}{%
\begin{tabular}{l cc cc}
\toprule
 & \multicolumn{2}{c}{ICC1k} & \multicolumn{2}{c}{ICC2k} \\
\cmidrule(lr){4-5}
\cmidrule(lr){2-3}
 & text & video & text & video \\
\midrule
Sentiment & 0.936 [0.930, 0.940] & 0.814 [0.770, 0.850] & 0.936 [0.930, 0.940] & 0.817 [0.760, 0.860] \\
Arousal & 0.635 [0.590, 0.670] & 0.802 [0.750, 0.840] & 0.666 [0.520, 0.760] & 0.807 [0.730, 0.860] \\
\bottomrule
\end{tabular}
}%

}

\end{table}%

\begin{table}

\caption{\label{tbl-cochrane_icc3k_testset}Inter-coder agreement in text transcript- and video-based ratings of valence/sentiment and arousal/intensity in subset of examples used for LLM scoring evaluation according to the average fixed raters intra-class correlation (ICCk3) coefficient.}

\centering{

\centering%
\normalsize%
\begin{tabular}{lcc}
\toprule
 & Text ratings & Video ratings \\
\midrule
Sentiment & 0.925 [0.910, 0.940] & 0.815 [0.730, 0.880] \\
Arousal & 0.686 [0.620, 0.740] & 0.839 [0.770, 0.890] \\
\bottomrule
\end{tabular}

}

\end{table}%

\begin{table}

\caption{\label{tbl-cochrane_kalphas}Inter-coder agreement in text transcript- and video-based ratings of valence/sentiment and arousal/intensity according to the Krippendorff's \(\alpha\) coefficient.}

\centering{

\centering%
\normalsize%
\begin{tabular}{lcc}
\toprule
 & Text ratings & Video ratings \\
\midrule
Sentiment & 0.828 & 0.556 \\
Arousal & 0.368 & 0.525 \\
\bottomrule
\end{tabular}

}

\end{table}%

\begin{figure}[!th]

\begin{minipage}{0.50\linewidth}

\centering{

\pandocbounded{\includegraphics[keepaspectratio]{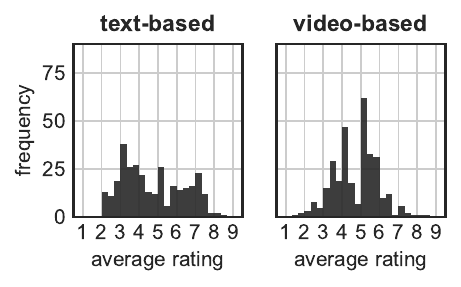}}

}

\subcaption{\label{fig-cochrane_outcomes_histograms_testset-1}Sentiment}

\end{minipage}%
\begin{minipage}{0.50\linewidth}

\centering{

\pandocbounded{\includegraphics[keepaspectratio]{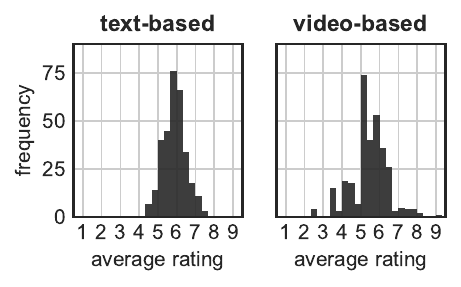}}

}

\subcaption{\label{fig-cochrane_outcomes_histograms_testset-2}Arousal}

\end{minipage}%

\caption{\label{fig-cochrane_outcomes_histograms_testset}Distribution of speech segment-level cross-coder averages of human ratings in the test set for text- and video-based sentiment/valence and arousal/activation annotations.}

\end{figure}%

\subsubsection{Cross-modality correlations of average human ratings}\label{cross-modality-correlations-of-average-human-ratings}

\begin{figure}[!th]

\centering{

\pandocbounded{\includegraphics[keepaspectratio]{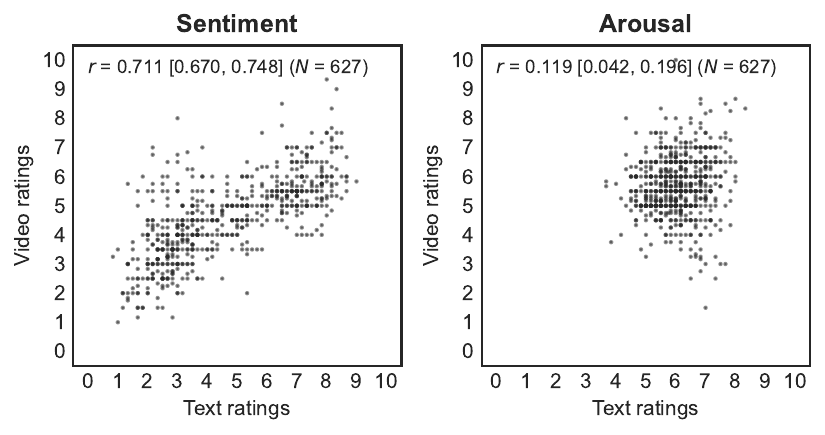}}

}

\caption{\label{fig-cochrane_crossmod_correlation}Correlation between speech segment-level averages of human coders' transcript (text) and video-based sentiment and arousal ratings. Measures compute by averaging annotators ratings within coding dimension and annotation modality.}

\end{figure}

Figure~\ref{fig-cochrane_crossmod_correlation} shows scatter plots and Pearson's correlation coefficients for sentiment and arousal measurements computed as average human ratings of speeches' videos and transcripts, respectively.
It shows that while for sentiment, average text- and video-based ratings are strongly positively correlated (Pearson's \(r = 0.711\)), for arousal, average text- and video-based ratings are only very weakly correlated (Pearson's \(r = 0.119\)).
This suggests that video- and text-based arousal ratings essentially measure different constructs.

\subsection{Test split descriptives}\label{test-split-descriptives}

\begin{figure}[!th]

\begin{minipage}{0.50\linewidth}

\centering{

\pandocbounded{\includegraphics[keepaspectratio]{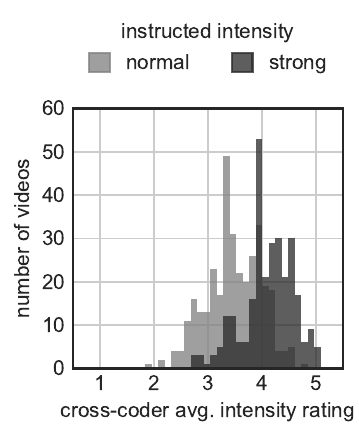}}

}

\subcaption{\label{fig-arousal_histograms_testsets-1}RAVDESS data. The shading of the histograms indicates whether actors were instructed to perform with normal or strong intensity.}

\end{minipage}%
\begin{minipage}{0.50\linewidth}

\centering{

\pandocbounded{\includegraphics[keepaspectratio]{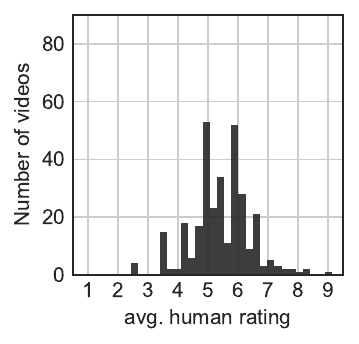}}

}

\subcaption{\label{fig-arousal_histograms_testsets-2}Parliamentary Speech Emotions data}

\end{minipage}%

\caption{\label{fig-arousal_histograms_testsets}Distribution of recording-level cross-coder averages emotional arousal/intensity ratings in the test splits of RAVDESS and Parliamentary Speech Emotions datasets.}

\end{figure}

Figure~\ref{fig-arousal_histograms_testsets} shows the distributions of the video recording-level average emotion intensity respectively arousal ratings in the test splits I use for mLLM evaluation.
For the RAVDESS data, I further group the data by the intensity level actors were instructed to apply when performing their lines.
On the 1--5 scale, average annotators' ratings center around 3.5 and 4.1 for recordings with normal and strong instructed intensity, respectively, while ratings are similarly dispersed in both conditions.
In the Cochrane et al. (2022) data, arousal scores are concentrated in the range between 4 and 6.5 on the 1--9 scale, indicating that most speech segments are rated as moderate in terms of arousal.
Yet, the distribution still covers most of the scale range, indicating that the data contains speech segments rated as low, moderate, and high in arousal.

\clearpage

\section{Prompts}\label{prompts}

\begin{prompt}[!h]
\caption{Prompt template for emotional intensity rating in RAVDESS data (Jinja2 prompt template)}\label{prompt:ravdess_arousal}
\begin{promptbox}
\begin{lstlisting}
Below you are presented with the {{modalities}} recording of a person speaking.

Your task is to rate the **intensity of the emotions expressed by the speaker** in the recording on a scale from {{low}} (low/weak) to {{high}} (high/strong).

## Clarifications

**Emotion intensity** refers to the degree or strength of emotional arousal, activation, or intensity in the speech. It captures whether the speech is animated or subdued, independently of whether it is positive or negative in sentiment.

## Your task

Rate the emotion intensity of the speech on a scale from {{low}} (low/weak) to {{high}} (high/strong) based on the {{modalities}} recording shown below.

- Return your rating as an integer value in the range {{low}} to {{high}}.
- Only return your rating and nothing else.

## Input ({{modalities}})
\end{lstlisting}
\end{promptbox}
\end{prompt}

\begin{prompt}[!h]
\caption{Prompt template for emotional arousal rating in Parliamentary Speech Emotions data (Jinja2 prompt template)}\label{prompt:cochrane_arousal}
\begin{promptbox}
\begin{lstlisting}
Below you are presented with the {{modalities}} of a short excerpt of a parliamentary speech.

Your task is to rate the level of **emotional arousal** of the speech on a scale from {{low}} (very subdue) to {{high}} (very aroused).

## Clarifications

**Emotional arousal** refers to the degree of emotional activation or intensity in the speech. It captures whether the speech is animated or subdued, independently of whether it is positive or negative in sentiment.

Emotional arousal varies from very subdue to very aroused.

- Very subdued (={{low}}) speech is calm, low-energy, or emotionally flat -- for example, conveying contentment or depression.
- Very aroused (={{high}}) speech is animated, intense, or highly expressive -- for example, expressing anger, excitement, or outrage.

## Your task

Rate the level of emotional arousal of the speech on a scale from {{low}} (very subdue) to {{high}} (very aroused) based on the speech's {{modalities}} shown below.
{% if multimodal %}
- Focus your analyses on the main speaker's speech and ignore background noises and movements.{% endif %}
- Return your rating as an integer value in the range {{low}} to {{high}}.
- Only return your rating and nothing else.

## Inputs ({{modalities}})
\end{lstlisting}
\end{promptbox}
\end{prompt}

\begin{prompt}[!h]
\caption{Prompt template for emotional valence (``sentiment'') rating in Parliamentary Speech Emotions data (Jinja2 prompt template)}\label{prompt:cochrane_valence}
\begin{promptbox}
\begin{lstlisting}
Below you are presented with the {{modalities}} of a short excerpt of a parliamentary speech.

Your task is to rate the **emotional valence** of the speech on a scale from {{low}} (very negative) to {{high}} (very positive).

## Clarifications

**Emotional valence** refers to the overall polarity of the emotions expressed in the speech -- that is, whether the speaker conveys a generally negative or positive emotional stance.

It varies from very negative to very positive.

- Very negative (={{low}}) speech conveys emotions such as anger, frustration, disappointment, or criticism.  
- Very positive (={{high}}) speech conveys emotions such as enthusiasm, approval, optimism, or praise.

## Your task

Rate the emotional valence of the speech on a scale from {{low}} (very negative) to {{high}} (very positive) based on the speech's {{modalities}} shown below.
{% if multimodal %}
- Focus your analyses on the main speaker's speech and ignore background noises and movements.{% endif %}
- Return your rating as an integer value in the range {{low}} to {{high}}.
- Only return your rating and nothing else.

## Inputs ({{modalities}})
\end{lstlisting}
\end{promptbox}
\end{prompt}

\clearpage

\section{Reproducibility}\label{sec-reproducibility}

Barrie, Palmer, and Spirling (2025) discuss how using large language models (LLMs) in political science research can complicate reproducibility and they propose best practices to address these challenges.
Below, I summarize how I have addressed these recommendations:

\begin{itemize}
\item
  \emph{Prefer and prioritize open-weight, locally versioned models:}
  I follow this recommendation by including the open-weights Qwen Omni and TowerVideo mLLMs in my analyses.
  In particular, I have recorded the concrete revisions of these models, allowing to download the exact model checkpoints used in my analyses.
\item
  \emph{Use closed/proprietary models only when justified:}
  I include the closed-weights Gemini Flash models to explore the current frontier of video-based in-context learning with mLLMs for emotion analysis.
\item
  \emph{Work in an ``anti-fragile'' way:}
  Generally, I rely on in-context learning.
  This means that a model's parameters are \emph{fixed} when generating response.
  This eliminates an important challenge to replicability: stochasticity arsing during deep neural network training on GPU hardware.

  Further, for open-weights models, I rely on local GPU inference with ``greedy decoding''.
  Relying on local GPU hardware means that I can control the computing environment (GPU instance(s), GPU driver version, python package version).
  Inference in ``greedy decoding'' mode means that at each text generation (``decoding'') step, the model selects the token with the highest generation probability as the next token in the response sequence.
  It is a \emph{deterministic} generation algorithm and parameters that govern randomness in generation (e.g., temperature, or top-\(k\) sampling) have no effect.
  In open-weights models, the only remaining source of nondeterminism is the GPU hardware and compute environment, which I have minimized by fixing the computation environment and batch (see Section~\ref{sec-reproducibility_impact}).
  In the case of the closed-weights Gemini Flash, the best one can do is to set the temperature parameter to 0 when calling the model through Google's Vertex AI API.

  An important limitation is inference with reasoning models (Gemini and Qwen 3 Omni Thinking) because ``thinking'' requires free-text generation. Here, setting temperature to 0 would reduce the quality of reasoning traces.
  I have thus used a default temperature of 1.0 when running reasoning models in reasoning mode (i.e., in ``thinking'' mode).
\item
  \emph{Replicate your own LLM annotations over time and report variance:}
  With open-weights models run on local hardware and greedy decoding, fixing the computation environment and batching should make computations deterministic.
  To assess the impact of the GPU hardware and computing environment (GPU driver and python versions), I discuss in Section~\ref{sec-reproducibility_impact} how an update to my local GPU setup in February 2026 has altered the mLLM scores I obtain.

  For the closed-weights Gemini Flash model, I observe an average standard deviation of less than 0.2 points on the 1--9 scale in a sample of examples in the Cochrane et al. (2022) data.
\item
  \emph{Provide uncertainty assessments for LM-generated labels:}
  I estimate confidence bounds on the performance metric estimates in my analysis using bootstrapping.
\end{itemize}

\subsection{Hardware configurations}\label{sec-gpu_configs}

Table~\ref{tbl-gpu_configs} reports the GPU configurations.
Single-GPU jobs ran on local hardware.
For multi-GPU setups, I ran models on the ETH Zurich's EULER cluster using SLURM.\footnote{see https://docs.hpc.ethz.ch/hardware/gpu\_nodes/ (accessed on \href{https://web.archive.org/web/20260401164011/https://docs.hpc.ethz.ch/hardware/gpu_nodes/}{April 1, 2026})}

\begin{table}

\caption{\label{tbl-gpu_configs}Hardware and configurations used for mLLM inference.}

\centering{

\centering%
\normalsize%

\resizebox{\textwidth}{!}{%
\begin{tabular}{rrrrr}
\toprule
 & Modalities & Shots & GPU Configuration & Quantization \\
Model &  &  &  &  \\
\midrule
Qwen 2.5 Omni 3B & video / text & 0, 3, and 5 shot & 1 $\times$ NVIDIA RTX 4090 (24 GB) & 4-bit \\
Qwen 2.5 Omni 7B & video / text & 0 and 3 shot & 1 $\times$ NVIDIA RTX 4090 (24 GB) & 4-bit \\
Qwen 2.5 Omni 7B & text & 5 shot & 1 $\times$ NVIDIA RTX 4090 (24 GB) & 4-bit \\
Qwen 2.5 Omni 7B & video & 5 shot & 2 $\times$ NVIDIA RTX 4090 (48 GB combined) & 4-bit \\
Qwen 3 Omni 30B A3B (Thinking / Instruct) & video / text  & 0 and 3 shot & 2 $\times$ NVIDIA RTX 4090 (48 GB combined) & 4-bit \\
Qwen 3 Omni 30B A3B (Thinking / Instruct) & text  & 5 shot & 2 $\times$ NVIDIA RTX 4090 (48 GB combined) & 4-bit \\
Qwen 3 Omni 30B A3B (Thinking / Instruct) & video & 5 shot & 2 $\times$ NVIDIA A100 (80 GB combined) & 4-bit \\
TowerVideo 2B & video & 0, 3, and 5 shot & 1 $\times$ NVIDIA RTX 4090 (24 GB) & 4-bit \\
TowerVideo 9B & video & 0 and 3 shot & 2 $\times$ NVIDIA RTX 4090 (48 GB combined) & 4-bit \\
\bottomrule
\end{tabular}

}%

}

\end{table}%

\subsection{Impact of GPU hardware and computing environment}\label{sec-reproducibility_impact}

I rely on local GPU hardware for in-context learning with open-weights mLLMs in my study.
This means that I can control the computing environment (GPU instance(s), GPU driver version, python package version).
Further, inference in ``greedy decoding'' mode means that at each text generation (``decoding'') step, the model selects the token with the highest generation probability as the next token in the response sequence.
It is a \emph{deterministic} generation algorithm and parameters that otherwise govern randomness in text generation (e.g., temperature, or top-\(k\) sampling) have no effect.

In open-weights models, the only remaining source of nondeterminism is hardware-level nondeterminism.
To assess the impact of the GPU hardware and computing environment (GPU driver and python versions), I leverage an update to my local GPU setup in February 2026.
This update allowed me to re-run the zero-shot video-based arousal scoring experiments in the Cochrane et al.~data with originally run with the hardware described in Table~\ref{tbl-gpu_configs} with a NVIDIA RTX PRO 6000 Blackwell Max-Q Workstation Edition.
Notably, this update to my local hardware involved increasing the version of \texttt{pytorch} (from 2.7.1 to 2.11.0), \texttt{torchvision} (from 2.7.1 to 2.11.0), \texttt{torchaudio} (from 0.22.1 to 0.25.0), and \texttt{transformers} (from 4.57.0 to 4.57.6).

Specifically, I re-ran these experiments with the Qwen 3 Omni Thinking, Qwen 3 Omni Instruct, and Qwen 2.5 Omni models because their distinct architectures.
In contrast to Qwen 2.5 Omni, Qwen 3 Omni models are Mixture-of-Experts (MoE) models.
As the routing of inference to a subset of ``experts'' in a MoE architecture adds a layer of non-determinism, changing the hardware and computing environment might affect these models scores more than for the Qwen 2.5 Omni model.

\begin{table}

\caption{\label{tbl-reproducibility_decomposition}Decomposition of deviations between original and reproduced scores into bias and noise components. The table shows the Pearson correlation, RMSE, MPE, and the proportions of bias and noise in the total error for all models.}

\centering{

\centering%
\normalsize%
\begin{tabular}{rrrrrr}
\toprule
 & Pearson's $r$ & RMSE$_{0–1}$ & MPE$_{0–1}$ & \\% bias & \\% noise \\
\midrule
qwen3-omni-thinking & 0.970 & 0.024 & -0.000 & 0.000 & 100.000 \\
qwen3-omni-instruct & 0.960 & 0.033 & 0.000 & 0.000 & 100.000 \\
qwen2.5-omni-7b & 0.996 & 0.013 & -0.000 & 0.100 & 99.900 \\
\bottomrule
\end{tabular}

}

\end{table}%

\begin{figure}[!th]

\centering{

\pandocbounded{\includegraphics[keepaspectratio]{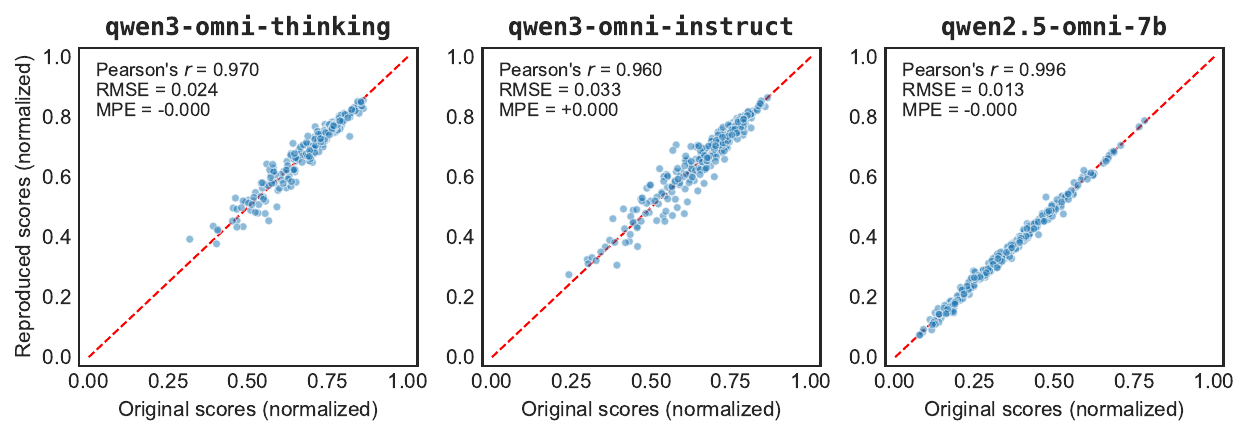}}

}

\caption{\label{fig-reproducibility_scatter}Scatter plots comparing original and reproduced scores for selected model. Figures plot models' normalized scores against each other, with a dashed red line indicating perfect agreement. Each plot includes annotations of Pearson's \(r\), RMSE, and MPE to quantify the agreement between original and reproduced scores.}

\end{figure}

Table~\ref{tbl-reproducibility_decomposition} and Figure~\ref{fig-reproducibility_scatter} show that, overall, the original and reproduced scores for the zero-shot video-based arousal scoring experiment in the test split of the Parliamentary Speech Emotions data are in high agreement.
This indicates a high degree of reproducibility even when changing the GPU hardware and computation environment (including GPU driver and python package versions).

\clearpage

\section{Results}\label{results-1}

\subsection{Scoring in the RAVDESS data}\label{scoring-in-the-ravdess-data}

\begin{landscape}

\begin{table}

\caption{\label{tbl-ravdess_scoring_res}Performance of mLLMs in video-based emotion intensity scoring in the RAVDESS data by model type and size and number of few-shot exemplars (0, 3, or 5). Performance measured in terms of correlation (ρ, \(r\)) and the root mean squared error (RMSE) evaluated against average scores in human coders' video-based ratings. Higher (lower) values in ρ and \(r\) (RMSE) indicate better performance. Reported values are averages of 120 bootstrapped estimates and their standard deviation. Column `Thinking' indicates whether the given reasoning model run in dynamic ``thinking'' mode or not. \emph{Note:} No results for TowerVideo-9B for 5-shot inference reported due to its context window size limitation.}

\centering{

\centering%
\normalsize%

\resizebox{1.3\textheight}{!}{%
\begin{tabular}{rr ccr ccr ccr}
\toprule
 &  & \multicolumn{3}{c}{Pearson's $r$} & \multicolumn{3}{c}{Spearman's ρ} & \multicolumn{3}{c}{RMSE$_{0–1}$} \\
\cmidrule(lr){9-11}
\cmidrule(lr){6-8}
\cmidrule(lr){3-5}
 &  & 0-shot & 3-shot & 5-shot & 0-shot & 3-shot & 5-shot & 0-shot & 3-shot & 5-shot \\
model & thinking &  &  &  &  &  &  &  &  &  \\
\midrule
\multirow[t]{2}{*}{gemini-3-flash-preview} & ``minimal'' & 0.668±0.022 & 0.671±0.021 & 0.669±0.023 & 0.654±0.021 & 0.657±0.020 & 0.653±0.022 & 0.283±0.009 & 0.205±0.006 & 0.179±0.005 \\
 & ``dynamic'' & 0.629±0.025 & 0.629±0.025 & 0.652±0.024 & 0.626±0.024 & 0.624±0.024 & 0.633±0.024 & 0.276±0.009 & 0.253±0.009 & 0.186±0.006 \\
\cmidrule(lr){1-11}
\multirow[t]{2}{*}{gemini-2.5-flash} & none & 0.690±0.021 & 0.667±0.023 & 0.662±0.022 & 0.682±0.020 & 0.651±0.023 & 0.646±0.022 & 0.196±0.006 & 0.186±0.006 & 0.200±0.006 \\
 & ``dynamic'' & 0.655±0.025 & 0.668±0.023 & 0.673±0.022 & 0.632±0.023 & 0.664±0.023 & 0.654±0.024 & 0.284±0.008 & 0.210±0.005 & 0.197±0.006 \\
\cmidrule(lr){1-11}
\multirow[t]{2}{*}{qwen3-omni-thinking} & none & 0.668±0.023 & 0.678±0.018 & 0.715±0.021 & 0.666±0.022 & 0.670±0.016 & 0.703±0.020 & 0.152±0.005 & 0.162±0.004 & 0.134±0.004 \\
 & ``dynamic'' & 0.626±0.025 & 0.649±0.021 & 0.636±0.024 & 0.598±0.024 & 0.623±0.020 & 0.592±0.025 & 0.322±0.008 & 0.244±0.007 & 0.215±0.006 \\
\cmidrule(lr){1-11}
qwen3-omni-instruct &   & 0.699±0.021 & 0.677±0.022 & 0.704±0.021 & 0.679±0.019 & 0.669±0.020 & 0.692±0.020 & 0.185±0.006 & 0.196±0.005 & 0.167±0.004 \\
\cmidrule(lr){1-11}
qwen2.5-omni-7b &   & 0.644±0.026 & 0.609±0.025 & 0.587±0.025 & 0.617±0.023 & 0.598±0.024 & 0.574±0.023 & 0.495±0.005 & 0.173±0.004 & 0.165±0.005 \\
\cmidrule(lr){1-11}
qwen2.5-omni-3b &   & 0.476±0.031 & 0.516±0.030 & 0.525±0.030 & 0.481±0.028 & 0.522±0.027 & 0.534±0.027 & 0.281±0.004 & 0.176±0.004 & 0.158±0.004 \\
\cmidrule(lr){1-11}
towervideo-9b &   & 0.484±0.029 & 0.401±0.030 & -/- & 0.431±0.028 & 0.388±0.028 & -/- & 0.260±0.004 & 0.233±0.005 & -/- \\
\cmidrule(lr){1-11}
towervideo-2b &   & 0.329±0.040 & 0.272±0.037 & 0.100±0.040 & 0.305±0.040 & 0.184±0.037 & 0.079±0.039 & 0.141±0.004 & 0.149±0.004 & 0.501±0.005 \\
\bottomrule
\end{tabular}
}%

}

\end{table}%

\end{landscape}

\begin{landscape}

\begin{table}

\caption{\label{tbl-ravdess_comp_thinking_scoring_res}Comparison of video-based arousal scoring results of Gemini 3 Flash Preview, Gemini 2.5 Flash, and Qwen 3 Omni Thinking in video recordings in the RAVDESS data test split depending on whether (dynamic) thinking was enabled or not (see `Thinking' indicator). Performance measured in terms of correlation (ρ, \(r\)) and the root mean squared error (RMSE) using averaged scores computed from human coders' ratings as a reference. Higher (lower) values in ρ and \(r\) (RMSE) indicate better performance. Error margin indicates standard deviation across 120 bootstrap resamples. Δ measures the relative performance difference (in percentage) when thinking mode is enabled compared to when it is disabled.}

\centering{

\centering%
\normalsize%

\resizebox{1.3\textheight}{!}{%
\begin{tabular}{rr ccr ccr ccr}
\toprule
 &   & \multicolumn{3}{c}{Pearson's $r$} & \multicolumn{3}{c}{Spearman's $\rho$} & \multicolumn{3}{c}{RMSE} \\
\cmidrule(lr){9-11}
\cmidrule(lr){6-8}
\cmidrule(lr){3-5}
 & \emph{Thinking} & no & yes & $\Delta$ & no & yes & $\Delta$ & no & yes & $\Delta$ \\
model & shots &  &  &  &  &  &  &  &  &  \\
\midrule
\multirow[t]{3}{*}{gemini-3-flash-preview} & 0-shot & 0.668$\pm$0.022 & 0.629$\pm$0.025 & -6\% & 0.654$\pm$0.021 & 0.626$\pm$0.024 & -4\% & 0.283$\pm$0.009 & 0.276$\pm$0.009 & +3\% \\
 & 3-shot & 0.671$\pm$0.021 & 0.629$\pm$0.025 & -6\% & 0.657$\pm$0.020 & 0.624$\pm$0.024 & -5\% & 0.205$\pm$0.006 & 0.253$\pm$0.009 & -19\% \\
 & 5-shot & 0.669$\pm$0.023 & 0.652$\pm$0.024 & -3\% & 0.653$\pm$0.022 & 0.633$\pm$0.024 & -3\% & 0.179$\pm$0.005 & 0.186$\pm$0.006 & -3\% \\
\cmidrule(lr){1-11}
\multirow[t]{3}{*}{gemini-2.5-flash} & 0-shot & 0.690$\pm$0.021 & 0.655$\pm$0.025 & -5\% & 0.682$\pm$0.020 & 0.632$\pm$0.023 & -7\% & 0.196$\pm$0.006 & 0.284$\pm$0.008 & -31\% \\
 & 3-shot & 0.667$\pm$0.023 & 0.668$\pm$0.023 & $\pm$0\% & 0.651$\pm$0.023 & 0.664$\pm$0.023 & +2\% & 0.186$\pm$0.006 & 0.210$\pm$0.005 & -11\% \\
 & 5-shot & 0.662$\pm$0.022 & 0.673$\pm$0.022 & +2\% & 0.646$\pm$0.022 & 0.654$\pm$0.024 & +1\% & 0.200$\pm$0.006 & 0.197$\pm$0.006 & +2\% \\
\cmidrule(lr){1-11}
\multirow[t]{3}{*}{qwen3-omni-thinking} & 0-shot & 0.668$\pm$0.023 & 0.626$\pm$0.025 & -6\% & 0.666$\pm$0.022 & 0.598$\pm$0.024 & -10\% & 0.152$\pm$0.005 & 0.322$\pm$0.008 & -53\% \\
 & 3-shot & 0.678$\pm$0.018 & 0.649$\pm$0.021 & -4\% & 0.670$\pm$0.016 & 0.623$\pm$0.020 & -7\% & 0.162$\pm$0.004 & 0.244$\pm$0.007 & -34\% \\
 & 5-shot & 0.715$\pm$0.021 & 0.636$\pm$0.024 & -11\% & 0.703$\pm$0.020 & 0.592$\pm$0.025 & -16\% & 0.134$\pm$0.004 & 0.215$\pm$0.006 & -38\% \\
\bottomrule
\end{tabular}
}%

}

\end{table}%

\end{landscape}

\begin{landscape}

\end{landscape}

\begin{figure}[!th]

\centering{

\pandocbounded{\includegraphics[keepaspectratio]{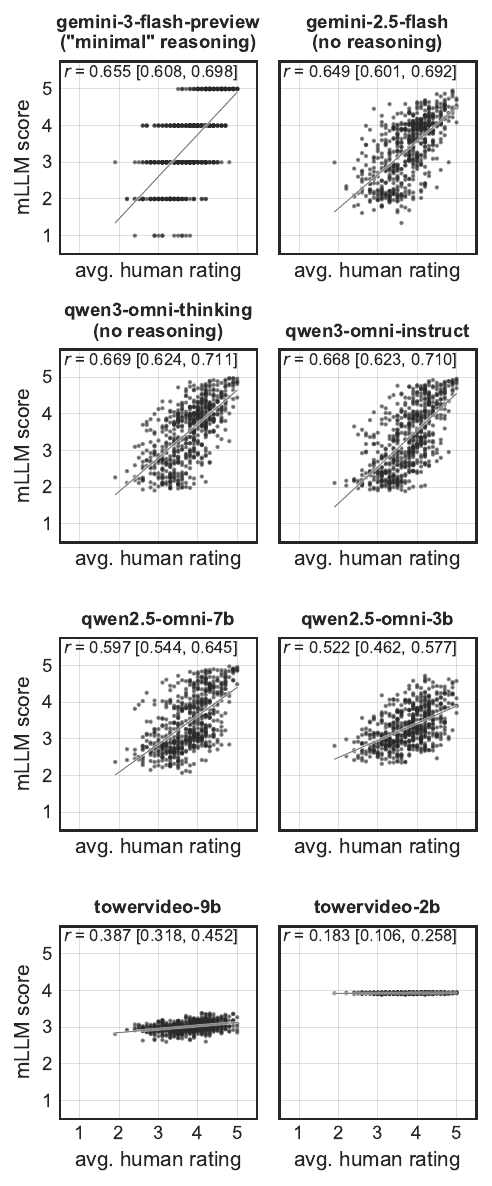}}

}

\caption{\label{fig-ravdess_scoring_scatter_all}Relationship between mLLM's video-based 3-shot emotion intensity scores and human coders' average ratings in the RAVDESS data test split. Correlation reported in terms of Pearson's \(r\). \emph{Note:} Correlation values reported here deviate from those in Figure~\ref{fig-ravdess_scoring_res} because the latter are bootstrapped average estimates instead of point estimates shown here.}

\end{figure}%

\begin{figure}[!th]

\centering{

\pandocbounded{\includegraphics[keepaspectratio]{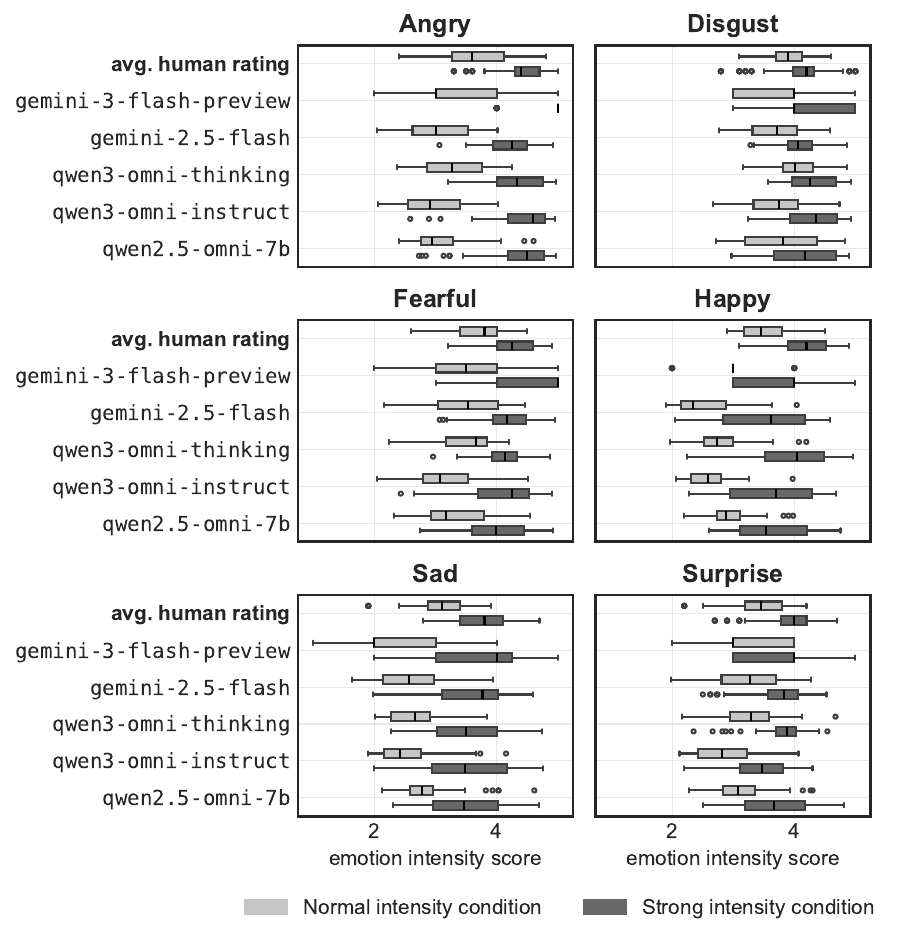}}

}

\caption{\label{fig-ravdess_scoring_boxplots}Distribution of video-based emotion intensity scores by emotion category, emotion intensity stimuli level, and scoring method (averaged human ratings or 3-shot mLLM inference).}

\end{figure}%

\clearpage

\subsubsection{Gender bias analysis}\label{sec-ravdess_gender_bias}

\begin{table}

\caption{\label{tbl-ravdess_error_decomposition_all}Error decomposition of mLLMs' video-based emotion intensity scores in the RAVDESS data test split, overall and by speaker gender. Numbers report Pearson correlation, root mean squared error (RMSE), and mean prediction error (MPE) measured on normalized 0--1 rating scale, and decomposition of scoring error into bias and noise components relative to overall variance.}

\centering{

\centering%
\normalsize%

\resizebox{\textwidth}{!}{%
\begin{tabular}{rrrrrrr}
\toprule
 &  & Pearson's $r$ & RMSE$_{0–1}$ & MPE$_{0–1}$ & \% bias & \% noise \\
model & group &  &  &  &  &  \\
\midrule
\multirow[t]{3}{*}{qwen3-omni-thinking} & \emph{overall} & 0.669 & 0.161 & -0.064 & 15.500 & 84.500 \\
 & group: female & 0.669 & 0.156 & -0.063 & 16.200 & 83.800 \\
 & group: male & 0.652 & 0.168 & -0.064 & 14.600 & 85.400 \\
\cmidrule(lr){1-7}
\multirow[t]{3}{*}{gemini-3-flash-preview} & \emph{overall} & 0.655 & 0.205 & -0.068 & 11.000 & 89.000 \\
 & group: female & 0.628 & 0.211 & -0.064 & 9.200 & 90.800 \\
 & group: male & 0.679 & 0.195 & -0.074 & 14.200 & 85.800 \\
\cmidrule(lr){1-7}
\multirow[t]{3}{*}{qwen2.5-omni-7b} & \emph{overall} & 0.597 & 0.172 & -0.079 & 20.800 & 79.200 \\
 & group: female & 0.569 & 0.168 & -0.058 & 12.100 & 87.900 \\
 & group: male & 0.592 & 0.178 & -0.107 & 35.900 & 64.100 \\
\cmidrule(lr){1-7}
\multirow[t]{3}{*}{gemini-2.5-flash} & \emph{overall} & 0.649 & 0.186 & -0.100 & 29.000 & 71.000 \\
 & group: female & 0.610 & 0.176 & -0.073 & 17.400 & 82.600 \\
 & group: male & 0.665 & 0.200 & -0.138 & 47.800 & 52.200 \\
\cmidrule(lr){1-7}
\multirow[t]{3}{*}{qwen3-omni-instruct} & \emph{overall} & 0.668 & 0.195 & -0.111 & 32.200 & 67.800 \\
 & group: female & 0.633 & 0.179 & -0.078 & 18.900 & 81.100 \\
 & group: male & 0.683 & 0.216 & -0.157 & 52.800 & 47.200 \\
\bottomrule
\end{tabular}

}%

}

\end{table}%

\paragraph{Potential bias in human ratings}\label{sec-ravdess_gender_bias_human_ratings}

To contextualize my findings regarding mLLMs' differential measurement error for male and femaler speakers in the RAVDESS test split, it is important to first understand potential gender bias in human ratings.
After all, I can only measure an mLLM's scoring error relative to the human rating-based reference scores because the ``true'' level of emotional intensity level with which an actor performed their line is unkown.

I assess potential gender bias in human raters' average ratings by regressing them on a speaker gender indicator and covariates that control for experimental condition (emotion, intensity, stimuli, statement, and repetition).
For comparaibility, I have normalized the human-rating average intensity scores to the 0--1 scale.
I then use three different modeling strategies to regress them on the speaker gender indicator and comntrols: a standard OLS regression, and OLS regression with standard errors clustered by speaker identity, and a mixed-effects modesl that treats observations as grouped within speakers.

\begin{table}

\caption{\label{tbl-rating_gender_bias_coefs}Estimated gender bias in huma raters' average emotion intensity ratings in the RAVDESS data test split based on linear regression modeling. Coefficient estimates reported for speaker gender indicator (reference: male) in three types of linear models: (1) OLS, (2) OLS with HC3 standard errors clustered by actor ID, and (3) mixed-effects model with random intercepts for actors. Statistical significance of the gender coefficient assessed through \(t\)-test in OLS model and \(z\)-test in mixed effects model. \emph{Note:} 95\% confidence intervals (CIs) reported.}

\centering{

\centering%
\normalsize%
\begin{tabular}{rrrrrr}
\toprule
 & $\beta_{gender}$ & Std. Err. & $p$-value & lower CI & upper CI \\
\midrule
OLS &     0.0551 &     0.009 &  0.000 &     0.038 &     0.072 \\
OLS w/ clustered SEs &     0.0551 &     0.022 &  0.013 &     0.011 &     0.099 \\
mixed effects & 0.055 & 0.025 & 0.025 & 0.007 & 0.103 \\
\bottomrule
\end{tabular}

}

\end{table}%

Table~\ref{tbl-ravdess_mllm_gender_bias_coefs} reports the results of a comparison of human raters' average intensity ratings of male and female actors recordings in the RAVDESS data relying on linear regression modeling shows that female actors arousal is rated about 0.055 points higher on the normalized 0--1 arousal scale than male actors' arousal when controlling for all observed experimental factors (emotion, intensity stimuli, statement, and repetition conditions).
This suggests a systematic upward bias in human raters' emotional intensity ratings of female speakers' recordings.
However, it is important to stress that since we do not know the ``true'' level of intensity with which actors performed their lines, the observed average arousal difference between female an male actors within experimental conditions could reflect that female actors indeed performed their lines with higher emotional intensity (e.g., due to socialization and gendered norms of emotional display).
In this case, the human raters would have reliably detected this difference.
Yet, as discussed in the main paper, I find bias in mLLMs' scores in the \emph{same} direction, so this possibility does not substantially alter my conclusion about the potentially compounding bias in mLLMs emoitional intensity ratings.

\paragraph{Alternative strategy to analyze differential bias in mLLM scores}\label{alternative-strategy-to-analyze-differential-bias-in-mllm-scores}

I examine the possibility of systematic differences in mLLM's systematic measurement error due to speaker gender in the RAVDESS data by decomposing theit scoring errors into bias and noise components and comparing the extent of systematic bias in videos of female and male speakers, respectively.

Below, I present the results of an alternative strategy that models the \emph{residuals} of mLLMs scores by regressing them on a speaker gender indicator and covariates that control for experimental condition (emotion, intensity, stimuli, statement, and repetition).
I compute the residuals of an mLLM's scores by subtracting them from the human-rating reference scores (both normalized to the 0--1 scale).
Use three different modeling strategies to regress them on the speaker gender indicator and comntrols: a standard OLS regression, and OLS regression with standard errors clustered by speaker identity, and a mixed-effects modesl that treats observations as grouped within speakers.

\begin{table}

\caption{\label{tbl-ravdess_mllm_gender_bias_coefs}Estimated gender bias in mLLMs' video-based emotion intensity scores in the RAVDESS data test split based on linear regression modeling of mLLMs' residuals. Coefficient estimates reported for speaker gender indicator (reference: male) in three types of linear models: (1) OLS, (2) OLS with HC3 standard errors clustered by actor ID, and (3) mixed-effects model with random intercepts for actors. Statistical significance of the gender coefficient assessed through \(t\)-test in OLS model and \(z\)-test in mixed effects model. \emph{Note:} 95\% confidence intervals (CIs) reported.}

\centering{

\centering%
\normalsize%

\resizebox{\textwidth}{!}{%
\begin{tabular}{rrrrrrr}
\toprule
 &  & $\beta_{gender}$ & Std. Err. & $p$-value & lower CI & upper CI \\
\midrule
\multirow[t]{3}{*}{qwen3-omni-thinking} & OLS &     0.0016 &     0.011 &  0.879 &    -0.019 &     0.022 \\
 & OLS w/ clustered SEs &     0.0016 &     0.034 &  0.962 &    -0.065 &     0.068 \\
 & mixed effects & -0.001 & 0.029 & 0.963 & -0.058 & 0.055 \\
\cmidrule(lr){1-7}
\multirow[t]{3}{*}{gemini-3-flash-preview} & OLS &     0.0096 &     0.013 &  0.464 &    -0.016 &     0.035 \\
 & OLS w/ clustered SEs &     0.0096 &     0.031 &  0.753 &    -0.050 &     0.069 \\
 & mixed effects & 0.008 & 0.031 & 0.801 & -0.053 & 0.068 \\
\cmidrule(lr){1-7}
\multirow[t]{3}{*}{qwen2.5-omni-7b} & OLS &     0.0484 &     0.012 &  0.000 &     0.025 &     0.072 \\
 & OLS w/ clustered SEs &     0.0484 &     0.037 &  0.185 &    -0.023 &     0.120 \\
 & mixed effects & 0.047 & 0.041 & 0.250 & -0.033 & 0.126 \\
\cmidrule(lr){1-7}
\multirow[t]{3}{*}{gemini-2.5-flash} & OLS &     0.0648 &     0.011 &  0.000 &     0.044 &     0.086 \\
 & OLS w/ clustered SEs &     0.0648 &     0.029 &  0.023 &     0.009 &     0.121 \\
 & mixed effects & 0.063 & 0.025 & 0.013 & 0.013 & 0.112 \\
\cmidrule(lr){1-7}
\multirow[t]{3}{*}{qwen3-omni-instruct} & OLS &     0.0796 &     0.011 &  0.000 &     0.058 &     0.101 \\
 & OLS w/ clustered SEs &     0.0796 &     0.033 &  0.015 &     0.016 &     0.144 \\
 & mixed effects & 0.077 & 0.033 & 0.019 & 0.013 & 0.142 \\
\bottomrule
\end{tabular}

}%

}

\end{table}%

Table~\ref{tbl-ravdess_mllm_gender_bias_coefs} reports the coefficient estimates, standard errors, p-values and 95\% confidence intervals for the speaker gender indicators from these regressions for all modeling appraoches and mLLMs examined in 3-shot inference.
The results are mostly in line with the estimates presented in Table~\ref{tbl-ravdess_bias_test_all}:
The usually positive coefficient estimate for the speaker gender indicators shows that, all else equal, mLLMs tend to rate female speakers' emotional intensity higher than that of male speakers.
Statistical significance tests, too, often support the conclusions reported in the main paper, with the exceptions that clustered standard errors and the mixed-effects models sometime indicate lack of statistical significance at conventional levels.

\clearpage

\subsubsection{Response confidence and token probability distributions}\label{response-confidence-and-token-probability-distributions}

\begin{figure}[!th]

\begin{minipage}{\linewidth}

\centering{

\pandocbounded{\includegraphics[keepaspectratio]{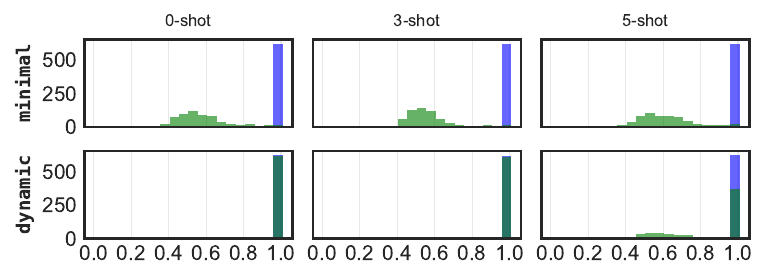}}

}

\subcaption{\label{fig-ravdess_response_token_probability_distributions_thinking_models-1}gemini-3-flash-preview}

\end{minipage}%
\newline
\begin{minipage}{\linewidth}

\centering{

\pandocbounded{\includegraphics[keepaspectratio]{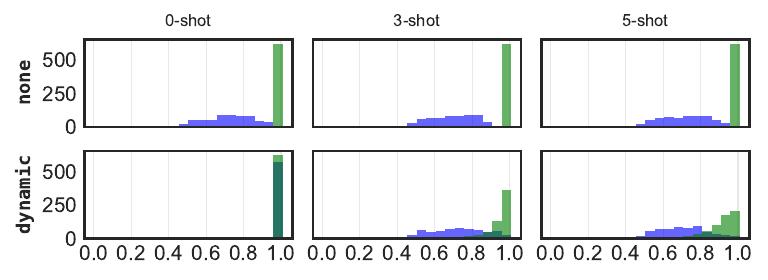}}

}

\subcaption{\label{fig-ravdess_response_token_probability_distributions_thinking_models-2}gemini-2.5-flash}

\end{minipage}%
\newline
\begin{minipage}{\linewidth}

\centering{

\pandocbounded{\includegraphics[keepaspectratio]{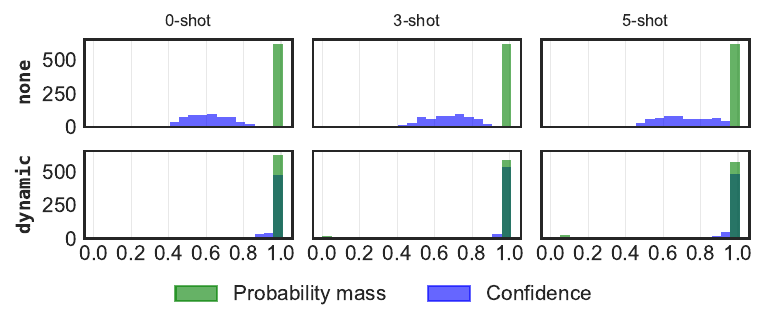}}

}

\subcaption{\label{fig-ravdess_response_token_probability_distributions_thinking_models-3}qwen3-omni-thinking}

\end{minipage}%

\caption{\label{fig-ravdess_response_token_probability_distributions_thinking_models}Distributions of response token probabilities assigned by different reasoning mLLMs when rating emotion intensity in the RAVDESS dataset depending on the amount of reasoning (plot panel rows) and the number of exemplars used for few-shot inference (columns). Histograms show the distribution of two metrics computed from the response token probabilities: Probability mass (green) measures the sum of the probabilities of all scale point option tokens (`1'-`5'); Confidence (blue) measures how much of the total scale point option token probability mass is assigned to the most likely scale point option token. The closer the confidence value distribution is to 1.0, the more the model is certain about its modal rating choice. The closer Probability mass value is to 1.0, the more the model conforms to the instruction to only respond with valid scale point options.}

\end{figure}%

Figure~\ref{fig-ravdess_response_token_probability_distributions_thinking_models} show the distribution of two metrics computed from the probabilities the mLLMs' assign to scale point option tokens for the different videos.
\emph{Probability mass} (green) measures the sum of the probabilities of all scale point option tokens (`1'-`5').
The closer the probability mass value is to 1.0, the more the model conforms to the instruction to only respond with a valid scale point option for a given video.
\emph{Confidence} (blue) measures how much of the total scale point option token probability mass is assigned to the most likely scale point option token.
The closer the confidence value for an video is to 1.0, the more certain is a model about its modal rating choice.

That confidence values are typically all (very close to) 1.0 when thinking is enabled (see blue histogram in bottom rows of plot panels) shows that enbaling thinking collapses the token probability distributions across all three reasoning mLLMs examined.
In comparison, when these models are configured to use only minimal (Gemini 3 Flash Preview) or no thinking at all (Gemini 2.5 Flash and Qwen 3 Omni Thinking), the token probability distributions are more spread out, leading to lower, and aruably more credible, confidence values.

Further, Figure~\ref{fig-ravdess_response_token_probability_distributions_thinking_models} shows that in the minimal thinking configuration, Gemini 3 Flash Preview struggles to conform to the instruction to only use valid scale point option tokens, as in many case a substantial share of its probability mass is spread across irrelevant vocabulary (green histogram distributed below 1.0).
This tendency also emerges for Gemini 2.5 Flash when reasoning is enabled in few-shot inference (see green histograms in bottom row, middle panel).

\begin{figure}[!th]

\centering{

\pandocbounded{\includegraphics[keepaspectratio]{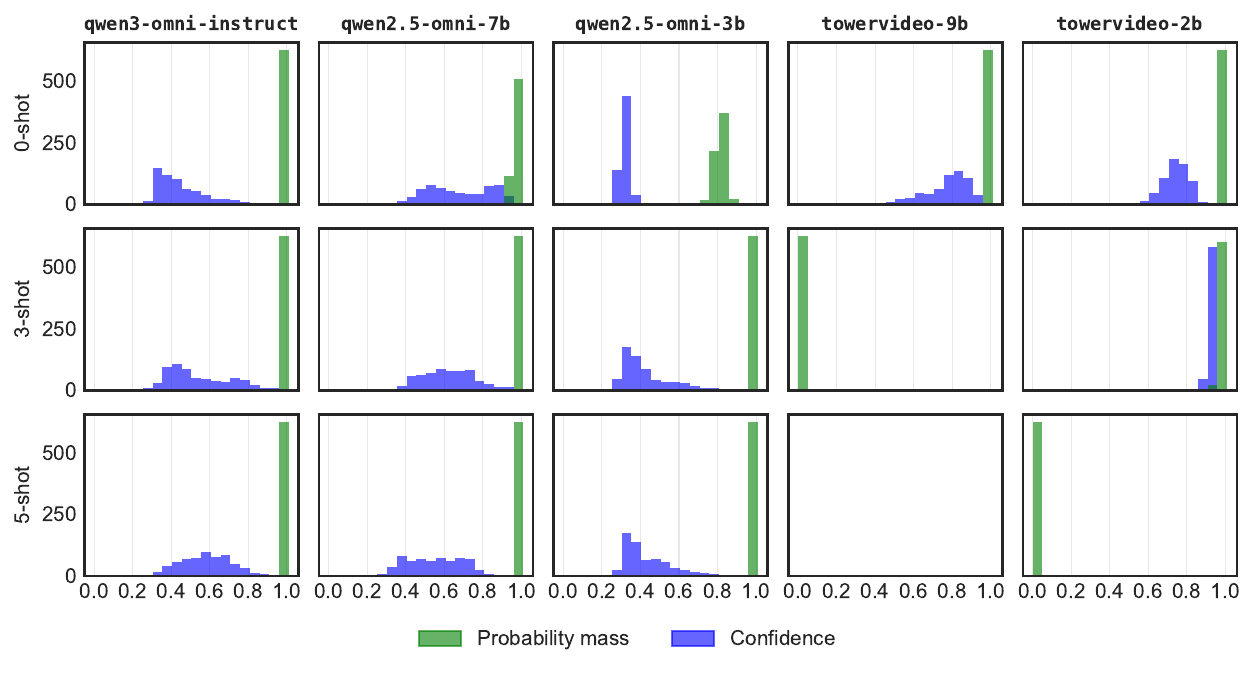}}

}

\caption{\label{fig-ravdess_response_token_probability_distributions_other_models}Distributions of response token probabilities assigned by different non-reasoning mLLMs when rating emotion intensity in the RAVDESS dataset depending the number of exemplars used in inference. Histograms show the distribution of two metrics computed from the response token probabilities: Probability mass (green) measures the sum of the probabilities of all scale point option tokens (`1'-`5'); Confidence (blue) measures how much of the total scale point option token probability mass is assigned to the most likely scale point option token. The closer the confidence value distribution is to 1.0, the more the model is certain about its modal rating choice. The closer Probability mass value is to 1.0, the more the model conforms to the instruction to only respond with valid scale point options.}

\end{figure}%

Figure~\ref{fig-ravdess_response_token_probability_distributions_other_models} presents the same analysis for non-reasoning mLLMs, including Qwen 3 Omni Instruct, Qwen 2.5 Omni (7B and 3B), and TowerVideo (9B and 2B).
Key observations include:

\begin{itemize}
\tightlist
\item
  Well-calibrated models show balanced confidence: Like qwen-3-omni-thinking without reasoning (see Figure~\ref{fig-ravdess_response_token_probability_distributions_thinking_models}), qwen-3-omni-instruct and qwen2.5-omni-7b demonstrate combine two important qualities: (1) most of their token probability mass (green bars, approaching 1.0) is allocated to valid response option tokens rather than being scattered across irrelevant vocabulary, and (2) their confidence in modal scale responses (blue bars) remains at reasonable levels rather than becoming extremely concentrated.
\item
  Model size and architecture impact confidence patterns: Comparing qwen-2.5-omni-3b to qwen-2.5-omni-7b reveals that larger models tend to exhibit better calibration, with higher probability mass assigned to valid response options and more balanced confidence distributions. This suggests that model size and architecture play a crucial role in shaping how mLLMs handle uncertainty in emotion intensity ratings.
\item
  TowerVideo models' very low performance in few-shot emotion intensity scoring in the RAVDESS data is explained by their failure to assign any meaningful probability mass to valid scale point option tokens (green bars close to 0.0).
\end{itemize}

\subsection{Scoring in the Parliamentary Speech Emotions data}\label{scoring-in-the-parliamentary-speech-emotions-data}

\begin{landscape}

\begin{table}

\caption{\label{tbl-cochrane_arousal_scoring_res}Performance of mLLMs in video-based arousal scoring of speech segment recordings in the Cochrane et al.~(2022) data by model type and model size and the number of few-shot exemplars (0, 3, or 5). Performance measured in terms of correlation (ρ, \(r\)) and the root mean squared error on the 0--1 scale(RMSE\(_{0-1}\)) using averaged scores computed from human coders' ratings as a reference. Higher (lower) values in ρ and \(r\) (RMSE\(_{0-1}\)) indicate better performance. Error margin indicates standard deviation across 120 bootstrap resamples. Column `Thinking' indicates whether the given reasoning model run in dynamic ``thinking'' mode or not. \emph{Note:} No results for TowerVideo-9B for 5-shot inference reported due to its context window size limitation.}

\centering{

\centering%
\normalsize%

\resizebox{1.3\textheight}{!}{%
\begin{tabular}{rr ccr ccr ccr}
\toprule
 &  & \multicolumn{3}{c}{Pearson's $r$} & \multicolumn{3}{c}{Spearman's ρ} & \multicolumn{3}{c}{RMSE$_{0–1}$} \\
\cmidrule(lr){9-11}
\cmidrule(lr){6-8}
\cmidrule(lr){3-5}
 &  & 0-shot & 3-shot & 5-shot & 0-shot & 3-shot & 5-shot & 0-shot & 3-shot & 5-shot \\
model & thinking &  &  &  &  &  &  &  &  &  \\
\midrule
\multirow[t]{2}{*}{gemini-3-flash-preview} & ``minimal'' & 0.373±0.050 & 0.431±0.052 & 0.362±0.054 & 0.330±0.050 & 0.385±0.054 & 0.347±0.057 & 0.191±0.007 & 0.159±0.007 & 0.143±0.007 \\
 & ``dynamic'' & 0.411±0.051 & 0.432±0.050 & 0.456±0.051 & 0.359±0.051 & 0.378±0.054 & 0.438±0.054 & 0.205±0.007 & 0.183±0.007 & 0.135±0.008 \\
\cmidrule(lr){1-11}
\multirow[t]{2}{*}{gemini-2.5-flash} & none & 0.411±0.046 & 0.430±0.046 & 0.430±0.048 & 0.334±0.048 & 0.392±0.051 & 0.383±0.051 & 0.166±0.007 & 0.166±0.006 & 0.152±0.007 \\
 & ``dynamic'' & 0.432±0.044 & 0.413±0.048 & 0.386±0.051 & 0.335±0.041 & 0.364±0.051 & 0.331±0.051 & 0.216±0.007 & 0.167±0.007 & 0.160±0.008 \\
\cmidrule(lr){1-11}
\multirow[t]{2}{*}{qwen3-omni-thinking} & none & 0.387±0.050 & 0.410±0.046 & 0.407±0.048 & 0.333±0.046 & 0.359±0.052 & 0.375±0.051 & 0.200±0.007 & 0.148±0.007 & 0.154±0.007 \\
 & ``dynamic'' & 0.337±0.057 & 0.380±0.050 & 0.407±0.050 & 0.255±0.049 & 0.345±0.055 & 0.355±0.054 & 0.258±0.009 & 0.170±0.008 & 0.161±0.007 \\
\cmidrule(lr){1-11}
qwen3-omni-instruct &   & 0.412±0.052 & 0.379±0.050 & 0.401±0.050 & 0.343±0.046 & 0.379±0.058 & 0.389±0.055 & 0.183±0.008 & 0.144±0.006 & 0.141±0.007 \\
\cmidrule(lr){1-11}
qwen2.5-omni-7b &   & 0.328±0.051 & 0.298±0.054 & 0.314±0.055 & 0.318±0.054 & 0.279±0.057 & 0.283±0.056 & 0.253±0.007 & 0.149±0.007 & 0.164±0.007 \\
\cmidrule(lr){1-11}
qwen2.5-omni-3b &   & 0.326±0.054 & 0.237±0.054 & 0.228±0.054 & 0.309±0.056 & 0.201±0.059 & 0.232±0.060 & 0.142±0.007 & 0.140±0.006 & 0.136±0.006 \\
\cmidrule(lr){1-11}
towervideo-9b &   & 0.258±0.050 & 0.201±0.058 & -/- & 0.226±0.047 & 0.231±0.062 & -/- & 0.174±0.006 & 0.123±0.006 & -/- \\
\cmidrule(lr){1-11}
towervideo-2b &   & 0.112±0.058 & 0.144±0.053 & -0.082±0.054 & 0.138±0.056 & 0.125±0.048 & -0.062±0.057 & 0.240±0.007 & 0.122±0.006 & 0.397±0.007 \\
\bottomrule
\end{tabular}
}%

}

\end{table}%

\end{landscape}

\begin{landscape}

\begin{table}

\caption{\label{tbl-cochrane_comp_thinking_scoring}Comparison of video-based arousal scoring results of Gemini 3 Flash Preview, Gemini 2.5 Flash, and Qwen 3 Omni Thinking in video recordings in the Parliamentary Speech Emotions data test split depending on whether (dynamic) thinking was enabled or not (see `Thinking' indicator). Performance measured in terms of correlation (ρ, \(r\)) and the root mean squared error on the 0--1 scale (RMSE\(_{0-1}\)) using averaged scores computed from human coders' ratings as a reference. Higher (lower) values in ρ and \(r\) (RMSE\(_{0-1}\)) indicate better performance. Error margin indicates standard deviation across 120 bootstrap resamples. Δ measures the relative performance difference (in percentage) when thinking mode is enabled compared to when it is disabled.}

\centering{

\centering%
\normalsize%

\resizebox{1.3\textheight}{!}{%
\begin{tabular}{rr ccr ccr ccr}
\toprule
 &   & \multicolumn{3}{c}{Pearson's $r$} & \multicolumn{3}{c}{Spearman's $\rho$} & \multicolumn{3}{c}{RMSE} \\
\cmidrule(lr){9-11}
\cmidrule(lr){6-8}
\cmidrule(lr){3-5}
 & \emph{Thinking} & no & yes & $\Delta$ & no & yes & $\Delta$ & no & yes & $\Delta$ \\
model & shots &  &  &  &  &  &  &  &  &  \\
\midrule
\multirow[t]{3}{*}{gemini-3-flash-preview} & 0-shot & 0.373$\pm$0.050 & 0.411$\pm$0.051 & +9\% & 0.330$\pm$0.050 & 0.359$\pm$0.051 & +8\% & 0.191$\pm$0.007 & 0.205$\pm$0.007 & -7\% \\
 & 3-shot & 0.431$\pm$0.052 & 0.432$\pm$0.050 & $\pm$0\% & 0.385$\pm$0.054 & 0.378$\pm$0.054 & -2\% & 0.159$\pm$0.007 & 0.183$\pm$0.007 & -13\% \\
 & 5-shot & 0.362$\pm$0.054 & 0.456$\pm$0.051 & +21\% & 0.347$\pm$0.057 & 0.438$\pm$0.054 & +21\% & 0.143$\pm$0.007 & 0.135$\pm$0.008 & +6\% \\
\cmidrule(lr){1-11}
\multirow[t]{3}{*}{gemini-2.5-flash} & 0-shot & 0.411$\pm$0.046 & 0.432$\pm$0.044 & +5\% & 0.334$\pm$0.048 & 0.335$\pm$0.041 & $\pm$0\% & 0.166$\pm$0.007 & 0.216$\pm$0.007 & -23\% \\
 & 3-shot & 0.430$\pm$0.046 & 0.413$\pm$0.048 & -4\% & 0.392$\pm$0.051 & 0.364$\pm$0.051 & -7\% & 0.166$\pm$0.006 & 0.167$\pm$0.007 & $\pm$0\% \\
 & 5-shot & 0.430$\pm$0.048 & 0.386$\pm$0.051 & -10\% & 0.383$\pm$0.051 & 0.331$\pm$0.051 & -14\% & 0.152$\pm$0.007 & 0.160$\pm$0.008 & -5\% \\
\cmidrule(lr){1-11}
\multirow[t]{3}{*}{qwen3-omni-thinking} & 0-shot & 0.387$\pm$0.050 & 0.337$\pm$0.057 & -13\% & 0.333$\pm$0.046 & 0.255$\pm$0.049 & -23\% & 0.200$\pm$0.007 & 0.258$\pm$0.009 & -22\% \\
 & 3-shot & 0.410$\pm$0.046 & 0.380$\pm$0.050 & -7\% & 0.359$\pm$0.052 & 0.345$\pm$0.055 & -4\% & 0.148$\pm$0.007 & 0.170$\pm$0.008 & -13\% \\
 & 5-shot & 0.407$\pm$0.048 & 0.407$\pm$0.050 & $\pm$0\% & 0.375$\pm$0.051 & 0.355$\pm$0.054 & -5\% & 0.154$\pm$0.007 & 0.161$\pm$0.007 & -5\% \\
\bottomrule
\end{tabular}
}%

}

\end{table}%

\end{landscape}

\begin{figure}[!th]

\centering{

\pandocbounded{\includegraphics[keepaspectratio]{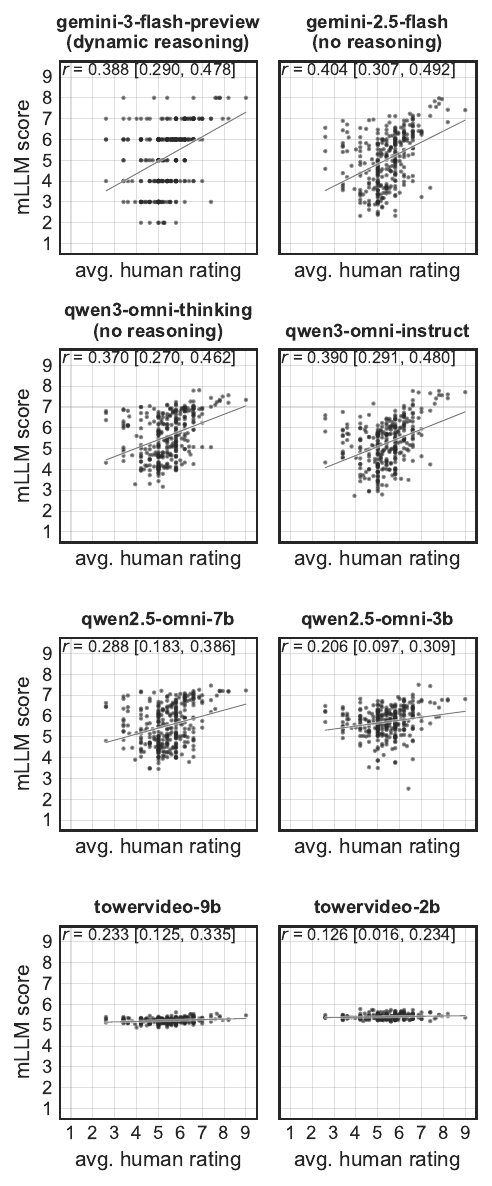}}

}

\caption{\label{fig-cochrane_arousal_scoring_scatter_plots_all}Scatter plots showing the relationship between human annotation-based reference scores and mLLMs' arousal scoring in 3-shot video-based in-context learning. The x-axis indicates the average human rating; the y-axis shows the score predicted by the mLLM. Correlation reported in terms of Pearson's \(r\). \emph{Note:} Correlation values reported here deviate from those in Figure~\ref{fig-cochrane_scoring_res} because the latter are bootstrapped average estimates in contrast to point estimates shown here.}

\end{figure}%

\begin{table}

\caption{\label{tbl-cochrane_error_decomposition_all}Error decomposition of mLLMs' video-based emotion intensity scores in the Parliamentary Speech Emotions data test split, overall and by speaker gender. Numbers report Pearson correlation, root mean squared error on the 0--1 scale (RMSE\(_{0-1}\)), and mean prediction error (MPE) measured on normalized 0--1 rating scale, and decomposition of scoring error into bias and noise components relative to overall variance.}

\centering{

\centering%
\normalsize%

\resizebox{\textwidth}{!}{%
\begin{tabular}{rrrrrrr}
\toprule
 &  & Pearson's $r$ & RMSE$_{0–1}$ & MPE$_{0–1}$ & \% bias & \% noise \\
\midrule
\multirow[t]{3}{*}{gemini-3-flash-preview} & \emph{overall} & 0.379 & 0.182 & -0.027 & 2.300 & 97.700 \\
 & group: female & 0.328 & 0.180 & -0.006 & 0.100 & 99.900 \\
 & group: male & 0.420 & 0.184 & -0.041 & 5.000 & 95.000 \\
\cmidrule(lr){1-7}
\multirow[t]{3}{*}{gemini-2.5-flash} & \emph{overall} & 0.403 & 0.166 & -0.049 & 8.800 & 91.200 \\
 & group: female & 0.365 & 0.172 & -0.028 & 2.700 & 97.300 \\
 & group: male & 0.439 & 0.162 & -0.063 & 15.000 & 85.000 \\
\cmidrule(lr){1-7}
\multirow[t]{3}{*}{qwen3-omni-thinking} & \emph{overall} & 0.354 & 0.148 & 0.025 & 2.800 & 97.200 \\
 & group: female & 0.283 & 0.158 & 0.044 & 7.700 & 92.300 \\
 & group: male & 0.408 & 0.142 & 0.012 & 0.800 & 99.200 \\
\cmidrule(lr){1-7}
\multirow[t]{3}{*}{qwen3-omni-instruct} & \emph{overall} & 0.376 & 0.143 & -0.018 & 1.500 & 98.500 \\
 & group: female & 0.339 & 0.141 & -0.001 & 0.000 & 100.000 \\
 & group: male & 0.409 & 0.144 & -0.029 & 3.900 & 96.100 \\
\cmidrule(lr){1-7}
\multirow[t]{3}{*}{qwen2.5-omni-7b} & \emph{overall} & 0.281 & 0.147 & 0.017 & 1.300 & 98.700 \\
 & group: female & 0.205 & 0.151 & -0.005 & 0.100 & 99.900 \\
 & group: male & 0.328 & 0.145 & 0.030 & 4.400 & 95.600 \\
\bottomrule
\end{tabular}

}%

}

\end{table}%

\clearpage

\subsubsection{Exploring the ``too much noise, too little signal'' explanation}\label{sec-apx_cochrane_explanation_1}

The promise of multimodality for computational analyses lies in adding nuance through acoustic and visual signals.
However, this potential added value can also be a challenge: acoustic and visual background noises and movements may dilute the relevant signal.

This problem of a lower signal-to-noise ratio may explain why the examined mLLMs are relatively reliable emotion intensity scorers in the ``clinically sterile'' RAVDESS data (see Section~\ref{sec-ravdess}) but fail to deliver on this promise in the more messy real-world parliamentary speech recordings in Cochrane et al.'s data.

Decomposing examined mLLMs' overal scoring errors in 3-shot inference in the Parliamentary Speech Emotions data into systematic and random error components (see Table~\ref{tbl-cochrane_error_decomposition_all}) substantiates this concern.
It shows that noise makes up 91.2\% or more of the deviations between mLLMs' scores and average human ratings.
For most models and few-shot configurations, this share (much) larger than in the RAVDESS data (see Table~\ref{tbl-ravdess_error_decomposition_all}).
Viewed together with the comparatively low correlation of mLLMs' with human-rating reference scores, this suggests that the examined mLLMs' ability to accurately rate speakers' arousal in this data might be hampered by nuisance factors particular to real-world video recordings.

What might be such diluting factors?
In the context of parliamentary debates, a nuisance in applications focusing on speakers' display of emotionality are ambient sounds in the form of background chatter, interjections, and interruptions.\footnote{This is likely espcially true for Westminster-style ``talking'' parliaments like the Canadian \emph{House of Commons} and Question Time sessions as those focused in Cochrane et al.~data.}
Further, in the Canadian \emph{House of Commons}, speakers often speak standing where they sit.
This means that in the background of the focussed speaker, there are often multiple other individuals, who might move and display various facial expressions.

Given the goal of analyzing the sentiment and level of emotional arousal in the \emph{speaker's} speech, reliable human annotators should ignore these factors and so should an mLLLM.
Accordingly, the relatively poor scoring performance of the mLLMs could be explained by their failure to discount accustic and visual background noises appropriately.

\begin{figure}[!th]

\begin{minipage}{0.50\linewidth}

\centering{

\pandocbounded{\includegraphics[keepaspectratio]{figures/figure_006.png}}

}

\subcaption{\label{fig-cochrane-masked-frame-original}original video frame}

\end{minipage}%
\begin{minipage}{0.50\linewidth}

\centering{

\pandocbounded{\includegraphics[keepaspectratio]{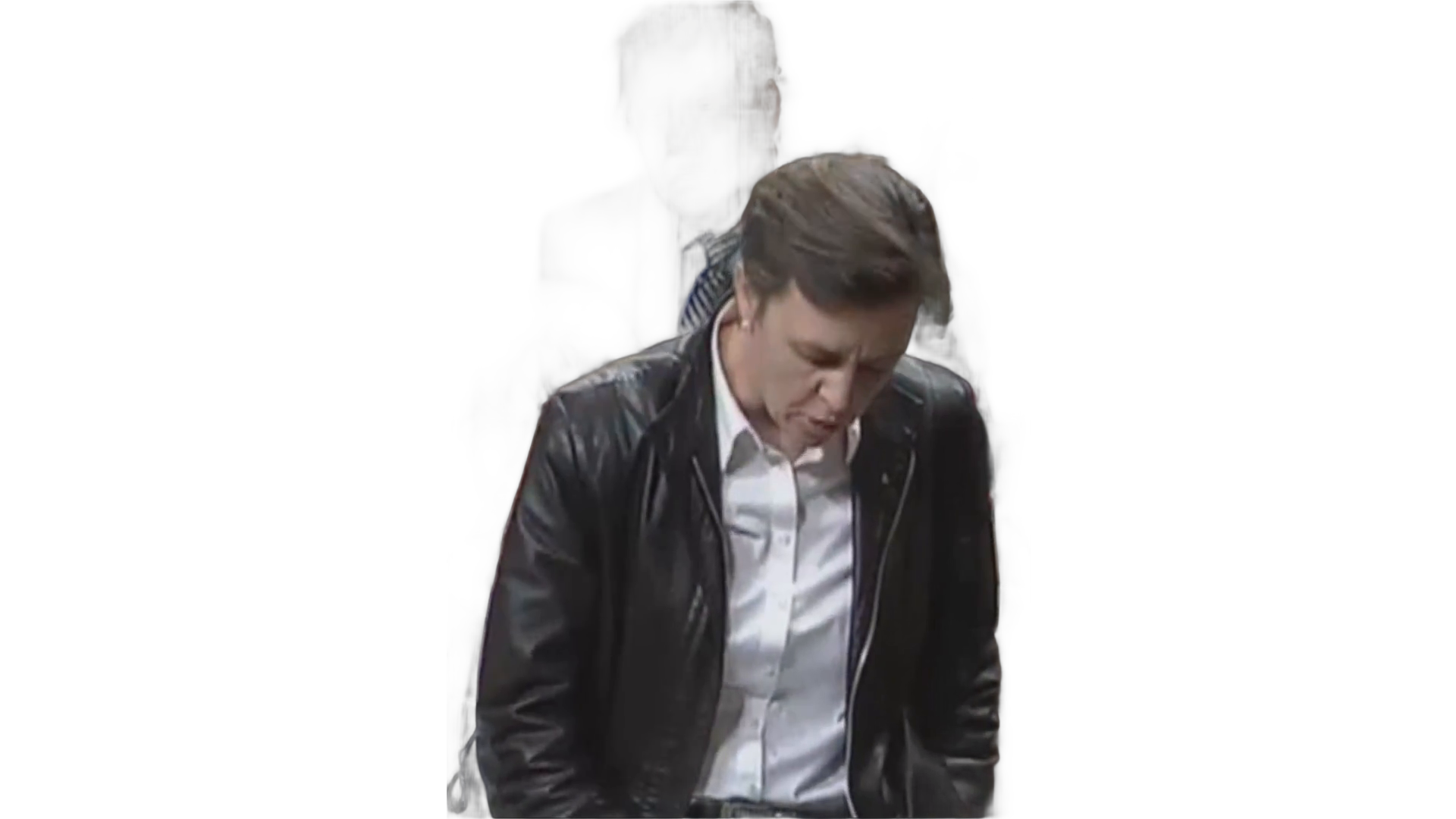}}

}

\subcaption{\label{fig-cochrane-masked-frame-masked}masked video frame}

\end{minipage}%

\caption{\label{fig-cochrane-masked-frame}Example of a video frame in the Cochrane et al.~(2022) data before and after masking background imagery.}

\end{figure}

To assess whether this failure to discount visual and acoustic background noises negatively impacts mLLMs' emotion scoring performance,
I have repeated the arousal scoring experiments reported above with a subset of 150 videos in the test set I have pre-processed to remove these nuisance factors as best as possible.
In particular, I have applied a noise filter to the videos' audio tracks that helps to focus a speaker's voice while suppressing background noises.
To remove visual background noise, I have applied a background masking strategy that relies on a pre-trained vLLM pre-trained for object detection that changes all pixels in a frame that are \emph{not} part of the speakers body to white.
This background image masking is illustrated in Figure~\ref{fig-cochrane-masked-frame}, which shows the same video frames from an example video before and after masking.

\begin{landscape}

\begin{table}

\caption{\label{tbl-cochrane_comp_cleaned_scoring_res}Comparison of video-based arousal scoring with selected mLLMs in original videos in the Cochrane et al.~(2022) data and their ``cleaned'' versions. The table compares models' performance in the same set of videos depending on whether acoustic and visual background noises have been `cleaned' from the videos, controlling for model size, resoning configuration (if applicable), and the number of few-shot in-context learning examples (0, 3, or 5). Performance measured in terms of correlation (ρ, \(r\)) and the root mean squared error on the 0--1 scale (RMSE\(_{0-1}\)) using averaged scores computed from human coders' ratings as a reference. Higher (lower) values in ρ and \(r\) (RMSE\(_{0-1}\)) indicate better performance. Δ measures the relative performance difference (in percentage) in arousal scoring in ``cleaned'' videos compared to their originals such that positive (negative) values indicate that in-context learning in cleaned (original) videos yields a better score on the given metric.}

\centering{

\centering%
\normalsize%

\resizebox{1.3\textheight}{!}{%
\begin{tabular}{rr ccr ccr ccr}
\toprule
 &  & \multicolumn{3}{c}{Pearson's $r$} & \multicolumn{3}{c}{Spearman's $\rho$} & \multicolumn{3}{c}{RMSE} \\
\cmidrule(lr){9-11}
\cmidrule(lr){6-8}
\cmidrule(lr){3-5}
 &  & original & cleaned & $\Delta$ & original & cleaned & $\Delta$ & original & cleaned & $\Delta$ \\
Model & Shots &  &  &  &  &  &  &  &  &  \\
\midrule
\multirow[t]{2}{*}{gemini-3-flash-preview} & 0-shot & 0.451$\pm$0.081 & 0.472$\pm$0.082 & +5\% & 0.428$\pm$0.077 & 0.403$\pm$0.085 & -6\% & 1.597$\pm$0.083 & 1.715$\pm$0.078 & -7\% \\
 & 3-shot & 0.493$\pm$0.080 & 0.448$\pm$0.077 & -9\% & 0.445$\pm$0.081 & 0.441$\pm$0.075 & -1\% & 1.428$\pm$0.086 & 1.409$\pm$0.086 & +1\% \\
\cmidrule(lr){1-11}
\multirow[t]{2}{*}{gemini-2.5-flash} & 0-shot & 0.490$\pm$0.067 & 0.480$\pm$0.069 & -2\% & 0.421$\pm$0.064 & 0.442$\pm$0.064 & +5\% & 1.287$\pm$0.083 & 1.274$\pm$0.079 & +1\% \\
 & 3-shot & 0.483$\pm$0.075 & 0.492$\pm$0.075 & +2\% & 0.448$\pm$0.074 & 0.458$\pm$0.077 & +2\% & 1.305$\pm$0.071 & 1.387$\pm$0.070 & -6\% \\
\cmidrule(lr){1-11}
\multirow[t]{2}{*}{qwen2.5-omni-7b} & 0-shot & 0.355$\pm$0.092 & 0.402$\pm$0.084 & +12\% & 0.331$\pm$0.090 & 0.383$\pm$0.087 & +14\% & 2.065$\pm$0.076 & 2.352$\pm$0.082 & -12\% \\
 & 3-shot & 0.469$\pm$0.074 & 0.511$\pm$0.080 & +8\% & 0.481$\pm$0.067 & 0.479$\pm$0.079 & $\pm$0\% & 0.951$\pm$0.058 & 1.073$\pm$0.087 & -11\% \\
\cmidrule(lr){1-11}
\multirow[t]{3}{*}{qwen2.5-omni-3b} & 0-shot & 0.416$\pm$0.079 & 0.348$\pm$0.081 & -16\% & 0.400$\pm$0.072 & 0.316$\pm$0.082 & -21\% & 1.098$\pm$0.083 & 1.013$\pm$0.080 & +8\% \\
 & 3-shot & 0.324$\pm$0.082 & 0.292$\pm$0.087 & -10\% & 0.260$\pm$0.089 & 0.249$\pm$0.080 & -4\% & 1.083$\pm$0.089 & 1.084$\pm$0.077 & $\pm$0\% \\
 & 5-shot & 0.325$\pm$0.075 & 0.256$\pm$0.081 & -21\% & 0.301$\pm$0.084 & 0.215$\pm$0.076 & -28\% & 1.044$\pm$0.079 & 1.170$\pm$0.074 & -11\% \\
\bottomrule
\end{tabular}
}%

}

\end{table}%

\end{landscape}

The results in Table~\ref{tbl-cochrane_comp_cleaned_scoring_res} provide no support for the hypothesis that visual and/or acoustic background noise hinder performance.
Most setups show no systematic improvements, suggesting additional factors beyond acoustic/visual noise explain poor real-world performance versus RAVDESS success.
Future research could examine whether body and head posture matters for mLLMs' video-based emotion analysis reliability.

\clearpage

\subsubsection{Comparison to results in sentiment scoring}\label{sec-apx_cochrane_sentiment_results}

To assess whether mLLMs' poor scoring performance in real-world parliamentary speech videos is specific to arousal measurement, I replicate the experiments reported above for sentiment scoring (see Prompt \ref{prompt:cochrane_valence}).
I focus on Gemini 2.5 Flash, Qwen 3 Omni Instruct, and Qwen 2.5 Omni, use the video-based sentiment ratings Cochrane et al. (2022) have collected as reference scores, and further use coders' transcript-based sentiment ratings to compare mLLMs' video-based to their text-based sentiment scoring performance.\footnote{The text vs.~video-based sentiment rating comparison is valid because, in contrast to arousal, the cross-coder averages of sentimnent ratings correlate strongly positively across annotation modalities (see Figure~\ref{fig-cochrane_crossmod_correlation}).
  For Qwen 2.5 Omni models, I use their Qwen 2.5 Instruct base LLMs as text-based reference in this analysis.}

\begin{figure}[!th]

\centering{

\pandocbounded{\includegraphics[keepaspectratio]{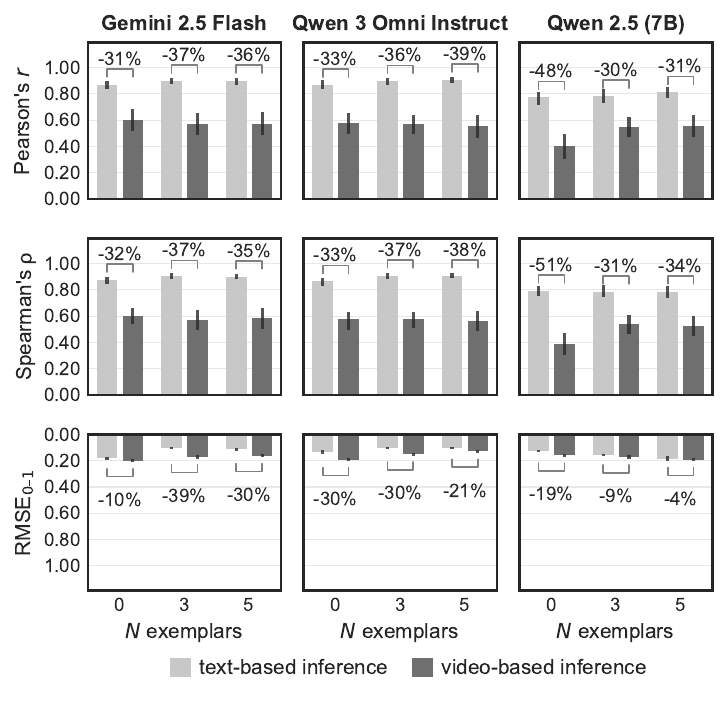}}

}

\caption{\label{fig-cochrane_sentiment_scoring_res}Comparison of video- and text-based sentiment scoring with Gemini 2.5 Flash, Qwen 3 Omni Instruct, and Qwen 2.5 models in video recordings and their transcripts, respectively, in the Parliamentary Speech Emotions data test split. Bars' height indicates performance measured in terms of correlation (ρ, \(r\)) and the root mean squared error on the 0-scale (RMSE\(_{0-1}\)) using averaged scores computed from human coders' ratings as a reference, where higher (lower) values in ρ and \(r\) (RMSE\(_{0-1}\)) indicate better performance. Bars put models' performances in text- and video-based inference side by side. Brackets plotted above pairs of bars report the relative (percentage point) performance difference of video-based in-context learning relative to text-based in-context learning such that negative values show that video-based in-context learning results in a worse score on the given metric. x-axis values indicate the number of few-shot in-context learning examples (0, 3, or 5).}

\end{figure}%

\begin{landscape}

\begin{table}

\caption{\label{tbl-cochrane_sentiment_scoring_res}Comparison of video- and text-based sentiment scoring with Qwen 2.5 Omni and Qwen 2.5 Instruct models in video recordings and their transcripts, respectively, contained in the test split of the Parliamentary Speech Emotions data test split. Performance measured in terms of correlation (ρ, \(r\)) and the root mean squared error on the 0--1 scale (RMSE\(_{0-1}\)) using averaged scores computed from human coders' ratings as a reference. Higher (lower) values in ρ and \(r\) (RMSE\(_{0-1}\)) indicate better performance. Error margin indicates standard deviation across 120 bootstrap resamples. Δ measures the relative performance difference (in percentage) in arousal scoring compared to sentiment scoring.}

\centering{

\centering%
\normalsize%

\resizebox{1.3\textheight}{!}{%
\begin{tabular}{rr ccr ccr ccr}
\toprule
 &  & \multicolumn{3}{c}{Pearson's $r$} & \multicolumn{3}{c}{Spearman's $\rho$} & \multicolumn{3}{c}{RMSE} \\
\cmidrule(lr){9-11}
\cmidrule(lr){6-8}
\cmidrule(lr){3-5}
 &  & text & video & $\Delta$ & text & video & $\Delta$ & text & video & $\Delta$ \\
Model & Shots &  &  &  &  &  &  &  &  &  \\
\midrule
\multirow[t]{3}{*}{Gemini 2.5 Flash} & 0-shot & 0.869$\pm$0.016 & 0.599$\pm$0.038 & -31\% & 0.875$\pm$0.016 & 0.599$\pm$0.032 & -32\% & 0.182$\pm$0.007 & 0.201$\pm$0.006 & -10\% \\
 & 3-shot & 0.898$\pm$0.013 & 0.570$\pm$0.046 & -37\% & 0.904$\pm$0.012 & 0.567$\pm$0.044 & -37\% & 0.103$\pm$0.005 & 0.170$\pm$0.008 & -39\% \\
 & 5-shot & 0.899$\pm$0.014 & 0.573$\pm$0.047 & -36\% & 0.900$\pm$0.012 & 0.583$\pm$0.044 & -35\% & 0.114$\pm$0.005 & 0.162$\pm$0.007 & -30\% \\
\cmidrule(lr){1-11}
\multirow[t]{3}{*}{Qwen 3 Omni Instruct} & 0-shot & 0.868$\pm$0.019 & 0.581$\pm$0.040 & -33\% & 0.864$\pm$0.017 & 0.575$\pm$0.037 & -33\% & 0.134$\pm$0.007 & 0.193$\pm$0.007 & -30\% \\
 & 3-shot & 0.898$\pm$0.014 & 0.573$\pm$0.042 & -36\% & 0.907$\pm$0.011 & 0.574$\pm$0.038 & -37\% & 0.105$\pm$0.005 & 0.150$\pm$0.006 & -30\% \\
 & 5-shot & 0.907$\pm$0.012 & 0.554$\pm$0.044 & -39\% & 0.908$\pm$0.010 & 0.566$\pm$0.041 & -38\% & 0.103$\pm$0.005 & 0.130$\pm$0.006 & -21\% \\
\cmidrule(lr){1-11}
\multirow[t]{3}{*}{Qwen 2.5 (7B)} & 0-shot & 0.774$\pm$0.024 & 0.405$\pm$0.050 & -48\% & 0.792$\pm$0.018 & 0.391$\pm$0.046 & -51\% & 0.128$\pm$0.004 & 0.158$\pm$0.006 & -19\% \\
 & 3-shot & 0.784$\pm$0.028 & 0.549$\pm$0.041 & -30\% & 0.785$\pm$0.023 & 0.540$\pm$0.040 & -31\% & 0.157$\pm$0.006 & 0.172$\pm$0.006 & -9\% \\
 & 5-shot & 0.812$\pm$0.024 & 0.558$\pm$0.042 & -31\% & 0.785$\pm$0.027 & 0.522$\pm$0.042 & -34\% & 0.184$\pm$0.010 & 0.192$\pm$0.006 & -4\% \\
\bottomrule
\end{tabular}
}%

}

\end{table}%

\end{landscape}

Figure~\ref{fig-cochrane_sentiment_scoring_res} shows mLLMs' video-based sentiment ratings align better with human scores than their arousal ratings (see Table~\ref{tbl-cochrane_sentiment_scoring_res}).
For example, in zero-shot inference, Gemini 2.5 Flash achieves a correlation of 0.599 (attenuation-adjusted: 0.657) and Qwen 2.5 Omni of 0.581 (0.637).
However, compared to the performance in text-based inference, video-based performance remains notably lower (holding constant model and few-shot setup).
For example, for Gemini 2.5 Flash, the ``reliability gap'' between video-based and text-based sentiment scoring (0.869 vs.~0.599 in zero shot-inference) is more than 2.5 times larger as the corresponding difference in human annotation (ICC3k 0.937 vs.~0.832).
This suggests systematic challenges in video-based inference persist beyond arousal scoring.

\clearpage

\subsubsection{Consequences in an illustrative downstream analysis}\label{sec-apx_cochrane_downstream_application}

Measuring a quantity of interest is often only the first step in applied research.
Thus, it is important to understand how substituting average human coders' arousal ratings with an mLLM's scores might affect substantive conclusions in downstream analyses.

I present an application example in the Parliamentary Speech Emotions data demonstrating that relying on mLLM's scores rather average human coders' arousal ratings might not necessarily change substantive conclusions in downstream analyses.
Importantly, this illustrative example is not intended as a systematic assessment of how relying on surrogate measurements obtained with machine learning methods can affect the conclusions drawn from downstream statistical inferences.
This question is at the heart of a growing literature (Baumann et al. 2025; Egami et al. 2024; TeBlunthuis, Hase, and Chan 2024; Knox, Lucas, and Cho 2022)and systematically addressing it in the context of multimodal emotion analysis is beyond the scope of this paper.
Rather, the application I present serves to demonstrates one concrete scenario where mLLM-generated scores could be deployed within a downstream analysis.

\begin{figure}[!th]

\centering{

\pandocbounded{\includegraphics[keepaspectratio]{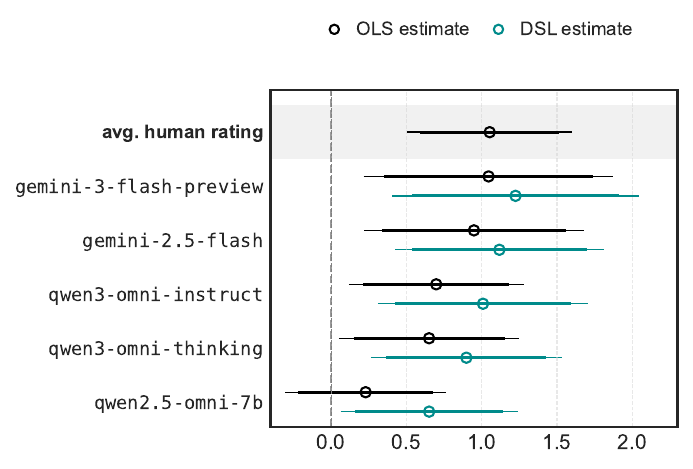}}

}

\caption{\label{fig-cochrane_gov_opposition_effects}Coefficient estimates of difference between opposition and government speakers' arousal in the Parliamentary Speech Emotions data test split when measured with avgerage human rating or mLLM scores, respectively. Coefficients estimates are shown on the x-axis and y-axis values indicate whether arousal was measures using avg. human rating or mLLM scores from video-based 3-shot in-context learning. Positive coefficient values indicate higher arousal scores for opposition speakers compared to government speakers. Colors indicate whether estimates were obtained with OLS regression (black) or using the design-bassed supervised learning method (cyan, Egami et al. 2024). Error bars indicate 90\% and 95\% confidence intervals around coefficients' point estimates. \emph{Notes:} All regressions adjusts for speakers' party affiliation, gender, and age group. DSL estimates computed using mLLMs scores as ``surrogate'' arousal measures and treating 75 (i.e., one third of) observations as having known ``ground truth'' measures.}

\end{figure}

My illustrative application examines the average difference in government and opposition speakers' arousal.
I estimate this difference using the videos in the Parliamentary Speech Emotions test split with an OLS regression that regresses a speeches arousal score on an indicator of the speaker's government or opposition status as well as indicators odf the speaker's gender, age group, and party affiliation.
The OLS coefficient estimate depicted in the top panel of Figure~\ref{fig-cochrane_gov_opposition_effects} shows that according to average human ratings, opposition speakers are significantly more aroused than government speakers when controlling for party, gender, and age (\(\hat{\beta} = 1.053\), \(\text{SE} = 0.28\), \(p < 0.001\); see Table~\ref{tbl-cochrane_regression}).
My application focuses on whether using the examined mLLMs' arousal scores as outcome in this regression leads to the same conclusion.
In addition to reporting OLS regression coefficient estimates, I apply the design-based supervised learning (DSL) method (Egami et al. 2024) to account for measurement error from mLLM outputs in this analysis.

Figure~\ref{fig-cochrane_gov_opposition_effects} shows that all mLLMs examined in 3-shot inference except Qwen 2.5 Omni (7B) produce arousal scores that lead to a substantively similar conclusion -- despite their at best moderate correlation with the human coding-based outcome measures.
This finding is strengthened when adopting the DSL method.
Using a small sample of 75 human-labeled examples to correct the OLS coefficient estimates for measurement error in the mLLM-based ``surrogate'' arousal scores, DSL tends to move the focal coefficient estimate closer to the reference estimate in fully human-labeled data.
This correction allows drawing a substantively very similar conclusion about the arousal difference between government and opposition speakers even when relying on the very poorly calibrated arousal scores obtained with Qwen 2.5 Omni (7B).

However, it is important to stress again that this application merely serves to demonstrates one concrete scenario where mLLM-generated scores could be deployed within a downstream analysis.
Whether insights derived from this specific example generalize to other downstream analyses remains an open question for future research.

\begin{landscape}

\begin{table}

\caption{\label{tbl-cochrane_regression}OLS regression results of regressing speaker arousal measures on speakers' opposition/government status, adjusting for gender, age group, and party affiliation. Columns present the results for regressions fitted using the respective measurements noted in column titles as outcomes in the regression.}

\centering{

 \centering
\resizebox{1.3\textheight}{!}{
\begin{tabular}{@{\extracolsep{5pt}}lcccccc}
\toprule
\\[-1.8ex] & \multicolumn{1}{c}{avg. human rating} & \multicolumn{1}{c}{\texttt{gemini-3-flash-preview}} & \multicolumn{1}{c}{\texttt{gemini-2.5-flash}} & \multicolumn{1}{c}{\texttt{qwen3-omni-thinking}} & \multicolumn{1}{c}{\texttt{qwen3-omni-instruct}} & \multicolumn{1}{c}{\texttt{qwen2.5-omni-7b}}  \\
\midrule
 Constant & 4.881$^{}$ & 5.199$^{}$ & 4.492$^{}$ & 6.174$^{}$ & 5.792$^{}$ & 5.573$^{}$ \\
& (0.985) & (1.487) & (1.308) & (1.075) & (1.041) & (0.956) \\
 Opposition speaker (\textit{ref.:} government) & 1.053$^{}$ & 1.046$^{}$ & 0.949$^{}$ & 0.651$^{}$ & 0.698$^{}$ & 0.230$^{}$ \\
& (0.280) & (0.423) & (0.372) & (0.306) & (0.296) & (0.272) \\
 Male speaker (\textit{ref.:} female) & 0.066$^{}$ & -0.244$^{}$ & -0.243$^{}$ & -0.169$^{}$ & -0.136$^{}$ & 0.368$^{}$ \\
& (0.111) & (0.167) & (0.147) & (0.121) & (0.117) & (0.108) \\
\multicolumn{7}{l}{Age group (\textit{ref.:} 24--44)} \\
 \quad 45--54 & -0.327$^{}$ & -0.756$^{}$ & -0.563$^{}$ & -0.612$^{}$ & -0.618$^{}$ & -0.626$^{}$ \\
& (0.130) & (0.197) & (0.173) & (0.142) & (0.138) & (0.127) \\
 \quad 55--79 & -0.341$^{}$ & -0.780$^{}$ & -0.535$^{}$ & -0.470$^{}$ & -0.525$^{}$ & -0.495$^{}$ \\
& (0.130) & (0.196) & (0.172) & (0.142) & (0.137) & (0.126) \\
 \midrule
 Party fixed-effects & Yes & Yes & Yes & Yes & Yes & Yes \\
\midrule
 Observations & 325 & 325 & 325 & 325 & 325 & 325 \\
 $R^2$ & 0.110 & 0.123 & 0.093 & 0.123 & 0.138 & 0.156 \\
 Adjusted $R^2$ & 0.082 & 0.095 & 0.064 & 0.095 & 0.110 & 0.129 \\
\bottomrule
\textit{Note:}\end{tabular}
}

}

\end{table}%

\end{landscape}

\begin{landscape}

\begin{table}

\caption{\label{tbl-cochrane_regression_dsl}DSL-adjusted OLS regression coefficients from regressing speaker arousal measures on speakers' opposition/government status, adjusting for gender, age group, and party affiliation. Columns present the results for regressions fitted using the respective measurements noted in column titles as outcomes in the regression. DSL (designe-based supervised learning) method implemented using a the average human arousal ratings of a random sample of 75 videos to adjust for measurement error in the mLLM-based arousal scoring.}

\centering{

 \centering
\resizebox{1.3\textheight}{!}{
\begin{tabular}{@{\extracolsep{5pt}}lccccc}
\toprule
& \multicolumn{5}{c}{\textit{Dependent variable: Emotional arousal score}} \
\cr \cline{2-6}
\\[-1.8ex] & \multicolumn{1}{c}{\texttt{gemini-3-flash-preview}} & \multicolumn{1}{c}{\texttt{gemini-2.5-flash}} & \multicolumn{1}{c}{\texttt{qwen3-omni-thinking}} & \multicolumn{1}{c}{\texttt{qwen3-omni-instruct}} & \multicolumn{1}{c}{\texttt{qwen2.5-omni-7b}}  \\
\midrule
 Constant & 3.195$^{}$ & 4.163$^{}$ & 4.114$^{}$ & -1.783$^{}$ & 4.582$^{}$ \\
& (1.882) & (1.008) & (1.251) & (6.970) & (0.690) \\
 Opposition speaker (\textit{ref.:} government) & 1.225$^{}$ & 1.118$^{}$ & 0.899$^{}$ & 1.009$^{}$ & 0.651$^{}$ \\
& (0.418) & (0.353) & (0.324) & (0.356) & (0.300) \\
 Male speaker (\textit{ref.:} female) & -0.254$^{}$ & -0.221$^{}$ & -0.220$^{}$ & -0.186$^{}$ & 0.146$^{}$ \\
& (0.160) & (0.146) & (0.121) & (0.118) & (0.109) \\
\multicolumn{6}{l}{Age group (\textit{ref.:} 24--44)} \\
 \quad 45--54 & -0.780$^{}$ & -0.621$^{}$ & -0.580$^{}$ & -0.650$^{}$ & -0.597$^{}$ \\
& (0.196) & (0.174) & (0.141) & (0.139) & (0.130) \\
 \quad 55--79 & -0.716$^{}$ & -0.444$^{}$ & -0.459$^{}$ & -0.531$^{}$ & -0.481$^{}$ \\
& (0.168) & (0.154) & (0.129) & (0.125) & (0.118) \\
 \midrule
 Party fixed-effects & Yes & Yes & Yes & Yes & Yes \\
\midrule
 Observations & 325 & 325 & 325 & 325 & 325 \\
\bottomrule
\textit{Note:}\end{tabular}
}

}

\end{table}%

\end{landscape}

\end{document}